\documentclass[sigconf]{acmart}
\AtBeginDocument{%
  \providecommand\BibTeX{{%
    \normalfont B\kern-0.5em{\scshape i\kern-0.25em b}\kern-0.8em\TeX}}}
\usepackage{graphicx,tabularx,subcaption}
\usepackage{booktabs}
\usepackage{multirow}
\usepackage[capitalise,nameinlink,compress]{cleveref}
\usepackage{xcolor}
\usepackage{comment}
\usepackage[english]{babel}
\usepackage{fontawesome5}
\usepackage{arydshln}
\usepackage{setspace} 
\usepackage{pdfpages}
\usepackage{framed}

\newenvironment{frshaded*}{%
\MakeFramed {\advance\hsize-\width \FrameRestore}
}%
{\endMakeFramed}

\graphicspath{{images/}}

\definecolor{main}{HTML}{5989cf}    % setting main color to be used
\definecolor{sub}{HTML}{cde4ff}     % setting sub color to be used

\copyrightyear{2024} 
\acmYear{2024} 
\setcopyright{acmlicensed}\acmConference[CHI '24]{Proceedings of the CHI Conference on Human Factors in Computing Systems}{May 11--16, 2024}{Honolulu, HI, USA}
\acmBooktitle{Proceedings of the CHI Conference on Human Factors in Computing Systems (CHI '24), May 11--16, 2024, Honolulu, HI, USA}
\acmDOI{10.1145/3613904.3642106}
\acmISBN{979-8-4007-0330-0/24/05}

\begin{document}


\title[EXMOS]{EXMOS: Explanatory Model Steering Through Multifaceted Explanations and Data Configurations}

%Directive explanations for monitoring the risk of diabetes onset: introducing data-centric explanations and combinations to support what-if explorations
\author{Aditya Bhattacharya}
\orcid{0000-0003-2740-039X}
\email{aditya.bhattacharya@kuleuven.be}
\affiliation{%
  \institution{KU Leuven}
  \city{Leuven}
  \country{Belgium}
}
\author{Simone Stumpf}
\orcid{0000-0001-6482-1973}
\email{Simone.Stumpf@glasgow.ac.uk}
\affiliation{%
  \institution{University of Glasgow}
  \city{Glasgow}
  \country{Scotland, UK}
}

\author{Lucija Gosak}
\orcid{0000-0002-8742-6594}
\email{lucija.gosak2@um.si}
\affiliation{%
  \institution{University of Maribor}
  \city{Maribor}
  \country{Slovenia}
}

% \authornote{These authors contributed equally.}
\author{Gregor Stiglic}
\orcid{0000-0002-0183-8679}
\email{gregor.stiglic@um.si}
\affiliation{%
  \institution{University of Maribor}
  \city{Maribor}
  \country{Slovenia}
}

\author{Katrien Verbert}
\orcid{0000-0001-6699-7710}
\email{katrien.verbert@kuleuven.be}
\affiliation{%
  \institution{KU Leuven}
  \city{Leuven}
  \country{Belgium}
}

\renewcommand{\shortauthors}{Bhattacharya, et al.}

\begin{abstract}
Explanations in interactive machine-learning systems facilitate debugging and improving prediction models. However, the effectiveness of various global model-centric and data-centric explanations in aiding domain experts to detect and resolve potential data issues for model improvement remains unexplored. This research investigates the influence of data-centric and model-centric global explanations in systems that support healthcare experts in optimising models through automated and manual data configurations. We conducted quantitative (n=70) and qualitative (n=30) studies with healthcare experts to explore the impact of different explanations on trust, understandability and model improvement. Our results reveal the insufficiency of global model-centric explanations for guiding users during data configuration. Although data-centric explanations enhanced understanding of post-configuration system changes, a hybrid fusion of both explanation types demonstrated the highest effectiveness. Based on our study results, we also present design implications for effective explanation-driven interactive machine-learning systems.
\end{abstract}


%\ccsdesc[500]{Human-centered computing~Visualization}%



\begin{teaserfigure}
  \centering
  \includegraphics[width=0.78\linewidth]{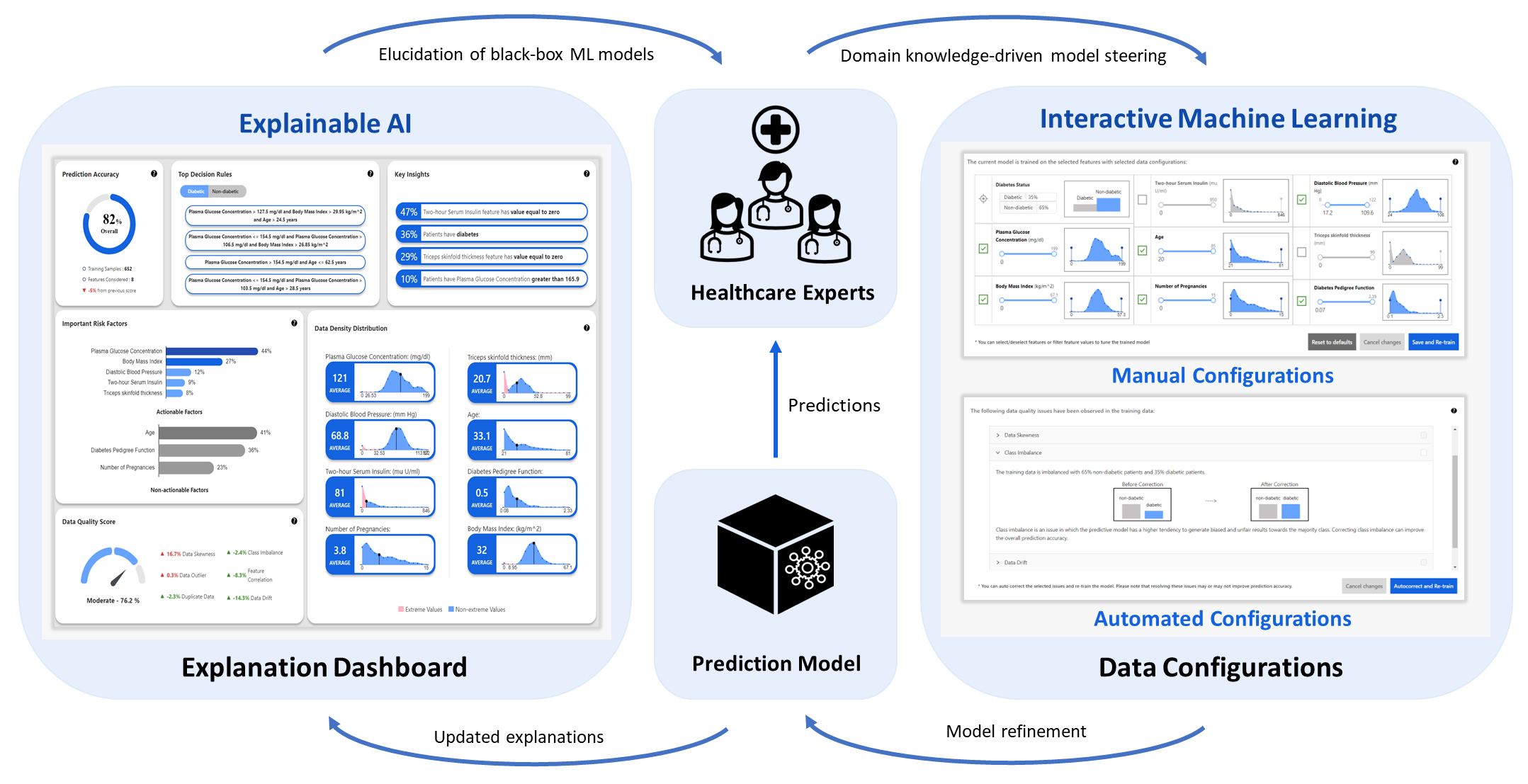}
  \caption{Explanatory Model Steering (EXMOS) enable users to fine-tune prediction models with the help of Explainable AI and Interactive Machine Learning. This research explores the influence of different types of global explanations for supporting domain experts, such as healthcare experts, in improving ML models through manual and automated data configurations. The refined prediction model also dynamically updates the explanations and the predicted outcomes.}
  \Description[Explanatory Model Steering for healthcare]{Explanatory Model Steering (EXMOS) enable users to fine-tune prediction models with the help of Explainable AI and Interactive Machine Learning. This research explores the influence of different types of global explanations for supporting domain experts, such as healthcare experts, in improving ML models through manual and automated data configurations. The refined prediction model also dynamically updates the explanations and the predicted outcomes.}
  \label{fig:xil_systems}
\end{teaserfigure}

\maketitle

\section{Introduction}
Encouraged by the promise of improved data-driven decision making, artificial intelligence (AI) and machine learning (ML) systems have witnessed increasing acceptance in high-stake domains such as healthcare~\cite{pawar2020incorporating, Caruana2015, Esteva2017-pg}. A pivotal aspect of these systems is the provision of explanations that enable end-users to develop a clearer mental model that fosters appropriate trust in the system, which is studied deeply in the field of Explainable AI (XAI)~\cite{Guidotti2018, adadi2018peeking}. 

Explanations are also useful for improving models through model steering~\cite{Wang_KDD_2022}. The field of Interactive Machine Learning (IML) studies approaches towards model steering and improvement, which are achieved by incorporating user feedback during ML model training~\cite{fails2003, Amershi_Cakmak_Knox_Kulesza_2014, teso2019}. Integrating explanations within IML systems enhances user understanding of ML models and fosters better interaction for model improvement~\cite{teso_leveraging_2022, teso2019, Schramowski2020, bertrand_chi_2023}. This emerging concept of explanatory model steering (EXMOS), which studies the joint effect of XAI and IML, has garnered attention as a deft human-centric solution to the challenges of acquiring rich end-user feedback for improving AI/ML systems~\cite{teso_leveraging_2022}. However, despite prior works demonstrating the positive impact of explanations on user understanding and control for model steering in IML systems~\cite{kulesza_explanatory_2010, kulesza_principles_2015}, the effectiveness
of different types of explanations~\cite{adadi2018peeking, BhattacharyaXAI2022, anik_data-centric_2021, hohman_gamut_2019} in model steering remains under-explored.

Furthermore, Schramowski et al.~\cite{Schramowski2020} have highlighted the importance of involving domain experts, i.e. users without ML backgrounds, in an explanation-driven model steering process, as domain knowledge is needed for identifying potentially misleading and biased predictors~\cite{feuerriegel2020fair}. Research has shown that current one-off explanations such as feature importances~\cite{ adadi2018peeking} or saliency maps~\cite{simonyan2014deep} are insufficient to support these users in model understanding ~\cite{lakkaraju2022rethinking}. Instead, domain experts require a better contextual understanding of the model through interactive explanations of the training data~\cite{lakkaraju2022rethinking}. With a better understanding of the training data, they can identify the limitations of the data and improve prediction models by configuring the training data.

While the value of explanations in IML systems is widely recognised, the specific impact of different types of global explanations for model steering by domain experts remains largely uncharted. Our research addresses this gap, investigating the effectiveness of different types of explanations, such as the global data-centric explanations~\cite{anik_data-centric_2021, BhattacharyaXAI2022}, model-centric explanations~\cite{adadi2018peeking, BhattacharyaXAI2022} and their combination within a healthcare-focused EXMOS system. Additionally, we explored data-centric approaches for model steering in which domain experts are involved to improve the quality of the training data~\cite{Mazumdar2022}. We investigated how different types of explanations motivate domain experts to improve prediction models when configuring the training data through two distinct approaches: (1) manual configuration and (2) automated configuration, as illustrated in \Cref{fig:manual_config} and \Cref{fig:auto_config}, respectively. The manual configuration approach enables domain experts, such as healthcare experts, to utilise their prior knowledge to assess the importance of predictor variables and mitigate bias or anomalies within the training data. In contrast, the automated configuration highlights potential issues in the training data~\cite{lones2023avoid, ackerman2022detection, kazerouni2020active} and allows users to select the issues that need correction. The system automatically applies correction algorithms to minimise these potential issues and retrains the
prediction model on the configured data.

Thus, this paper probes into the following research questions:

\begin{description}
\item[RQ1.] How do different types of global explanations affect healthcare experts’ ability to configure training data and enhance the prediction model’s performance, and why?
\item[RQ2.] To what extent do different types of global explanations influence healthcare experts’ trust and understanding of the AI system?
\item[RQ3.] How do different types of global explanations impact the choice of steering models through training data configuration?
\end{description}

To address these questions, we first developed a prototype EXMOS system (as illustrated in \Cref{fig:xil_systems}) with three versions of the explanation dashboard: (1) a Data-Centric Explanation version that included only global data-centric explanations, (2) a Model-Centric Explanation version that included only global model-centric explanations, and (3) a Hybrid version that combined all the explanations from data-centric and model-centric versions to provide multifaceted explanations~\cite{BhattacharyaXAI2022, wang_designing_2019, hohman_gamut_2019}. Then, to investigate the influence of different explanation dashboards on trust, understanding and model improvement, we conducted a between-subject quantitative study and another between-subject qualitative study involving 70 and 30 healthcare experts, respectively. 

Results indicate that the hybrid version participants were significantly better in prediction model improvement despite having a higher perceived task load than data-centric and model-centric participants. However, elevated perceived task load did not negatively impact their understanding or trust of the system. Findings also indicate the limitations of global model-centric explanations for guiding users during data configuration. Global data-centric explanations were particularly helpful when understanding post-configuration system changes as these provide more holistic elucidation of the training data. 

To summarise, there are three primary research contributions presented in this paper:
\begin{description}
\item [1.] We instantiated generic designs of global data-centric explanations, model-centric explanations, and a hybrid version that combined these different explanation types through our healthcare-focused EXMOS system. We propose a set of visualisation and interaction designs of explanations and data configurations to facilitate domain experts in model steering. We have open-sourced our system on \anon[GitHub]{GitHub}\footnote{\url{https://github.com/adib0073/EXMOS/}}. 
\item [2.] We evaluated the impact of these different types of explanations in model steering by domain experts through two extensive user studies involving healthcare experts. Our findings indicate that the hybrid combination of global explanations proved the most effective and efficient for steering models.
\item [3.] Based on the results of these user studies, we present guidelines for designing explanations and data configuration mechanisms to facilitate domain experts in model steering. 
\end{description}

\section{Background and related work}
To contextualise our research, this section discusses prior work on various XAI methods for ML systems and different model steering approaches in an IML workflow.

\subsection{XAI Methods for ML Systems}
Over the past decade, the field of XAI has witnessed numerous studies being conducted to measure the efficacy of various explanation methods for increasing the transparency of ML systems across multiple application domains, such as healthcare~\cite{pawar2020incorporating, Bhattacharya2023, Caruana2015, Che2017}, finance~\cite{bove_contextualization_2022, Bussmann2021, Fahner2018}, and law enforcement~\cite{soares2019fairbydesign, wang2022pursuit, Zeng2016}. Along with making black-box ML models more transparent, XAI methods have also aimed to make these systems more understandable and trustworthy ~\cite{Miller2017, liao2022connecting, Bhattacharya2023}.

Explanation methods have been categorised as model-specific or model-agnostic based on the degree of specificity~\cite{adadi2018peeking, anik_data-centric_2021}. Methods that can be applied to only specific model architectures and algorithms are termed model-specific explanations, like Saliency Maps~\cite{simonyan2014deep}, Grad-CAMs~\cite{Selvaraju_2019}, Visual Attention Maps~\cite{hassanin2022visual} etc. On the contrary, model-agnostic explanations can explain any model irrespective of the algorithms used. Popular XAI methods such as LIME~\cite{ribeiro2016why}, SHAP~\cite{lundberg2017unified}, and Surrogate Explainers~\cite{hepburn2021explainers} are examples of model-agnostic methods.

Based on the scope of explanations, XAI methods are also categorised as local explanations and global explanations~\cite{adadi2018peeking}. Local explanations involve explaining individual predictions considering a specific record, whereas global explanations describe the whole model trained on the entire dataset. Prior works have shown that global explanations induce more confidence in understanding the model compared to local explanations~\cite{anik_data-centric_2021, Dodge_2019, Popordanoska2020MachineGH}.

Based on the dimensions of explainability~\cite{BhattacharyaXAI2022, wang_designing_2019, hohman_gamut_2019}, researchers have further classified explanations as model-centric and data-centric. Model-centric methods such as SHAP-based feature importance explanations~\cite{lundberg2017unified, adadi2018peeking} aim to estimate the importance of parameters and hyper-parameters used in ML models. Data-centric explanations, on the other hand, aim to find insights from the training data to justify the behaviour of prediction models~\cite{anik_data-centric_2021}. Recent works have shown that data-centric explanations can justify the failure of ML models by revealing bias, inconsistencies and quality of the training data~\cite{anik_data-centric_2021, Bhattacharya2023, BhattacharyaXAI2022}. Examples of data-centric explanation approaches include summarisation of the training data using descriptive statistics, disclosing the bias in training data by showing the distribution of the data across various demographic parameters and revealing the potential issues that can impact the data quality~\cite{anik_data-centric_2021, Bhattacharya2023, BhattacharyaXAI2022}. 

Previous studies have shown that local data-centric explanations are more effective than local model-centric explanations for increasing the trust and understandability of prediction models by justifying predicted outcomes with reference to the training data~\cite{demsar2019, Bhattacharya2023}. However, these studies have focused on the efficacy of these explanations solely in the context of prediction justification of individual data instances (i.e., local explanations) rather than the working of the whole model (i.e., global explanations). Moreover, these studies have been conducted only with traditional ML systems, which do not consider user feedback for model steering. Consequently, the effectiveness of global data-centric explanations, model-centric explanations and their combinations for supporting domain experts to steer models remain unexplored in existing research. Our work aims to address this gap by investigating the importance of these different types of model-agnostic global explanations and their combinations for a healthcare-focused system.

\subsection{Model Steering Approaches in IML Systems}
The study of collaborative user interactions with ML systems that guide users in rectifying erroneous predictions is gaining increased attention~\cite{Amershi_Cakmak_Knox_Kulesza_2014, kulesza_explanatory_2010, kulesza_principles_2015, Krause2016InteractingWP, Schramowski2020, spinner2019explainer}. The term \textit{interactive machine learning} was introduced in Fails et al.’s work, which described the usage of a train-feedback-correct cycle involving end-users to rectify mistakes in an image segmentation system~\cite{fails2003}. Since then, extensive work has been conducted to support human-in-the-loop approaches by engaging end-users in model development, evaluation and correction for optimising ML systems~\cite{muralidhar2018incorporating, spinner2019explainer, stumpf2009interacting, Guo2022BuildingTI, honeycutt2020soliciting}.

Popular approaches include better model selection by end-users \cite{Amershi2010, kulesza_principles_2015, Fiebrink2011, Talbot2009}, elicitation of labels for important instances during active learning~\cite{kulesza_principles_2015, cakmak2011mixed,Settles2009ActiveLL}, improvement of reinforcement learning process for automated agents~\cite{knox2012}. In mixed-initiative active learning~\cite{settles2011closing}, both the end-user and the ML model share the responsibility of selecting the instances. Researchers have also proposed other approaches that enable users to inspect model parameters, such as features and their weights~\cite{cho2019explanatory} or rules used to make decisions~\cite{yang2019study} and enable them to modify these parameters. Moreover, the use of visual interfaces in the IML workflow has been emphasised to increase end-user involvement~\cite{sacha2018vis4ml}.
 
In recent work on \textit{explanatory interactive learning}, the model predictions are explained to the user as a basis for them to improve explanation reasoning~\cite{teso2019}. We found this approach promising and wanted to investigate further how explanatory interactive learning can enable domain experts to improve prediction models.

However, AI researchers have recently highlighted the need for data-centric AI  by stressing the importance of good quality data for more reliable prediction models~\cite{Mazumdar2022, jarrahi2022principles}. Conventional model-centric AI focuses on improving AI performance through better model selection and fine-tuning of hyperparameters, thus neglecting potential data quality issues and flaws in the training data ~\cite{Whang2020, zha2023datacentric, jarrahi2022principles, BhattacharyaXAI2022, brown2020language}. 

Particularly in high-stake domains, such as healthcare, high-quality training data is even more critical for the increased adoption of AI systems~\cite{chang2023knowledgeguided}. Hence, the active involvement of domain experts is crucial for identifying and correcting training data issues for model improvement. Yet, the involvement of domain experts in model steering following data-centric approaches is under-explored in the literature. To address this gap, our research investigated how healthcare experts can improve the flaws in the training data for better prediction models by applying their domain knowledge through manual and automated data configurations.

\begin{figure*}
\centering
\begin{subfigure}[b]{0.78\textwidth}
\centering
\includegraphics[width=1.0\linewidth]{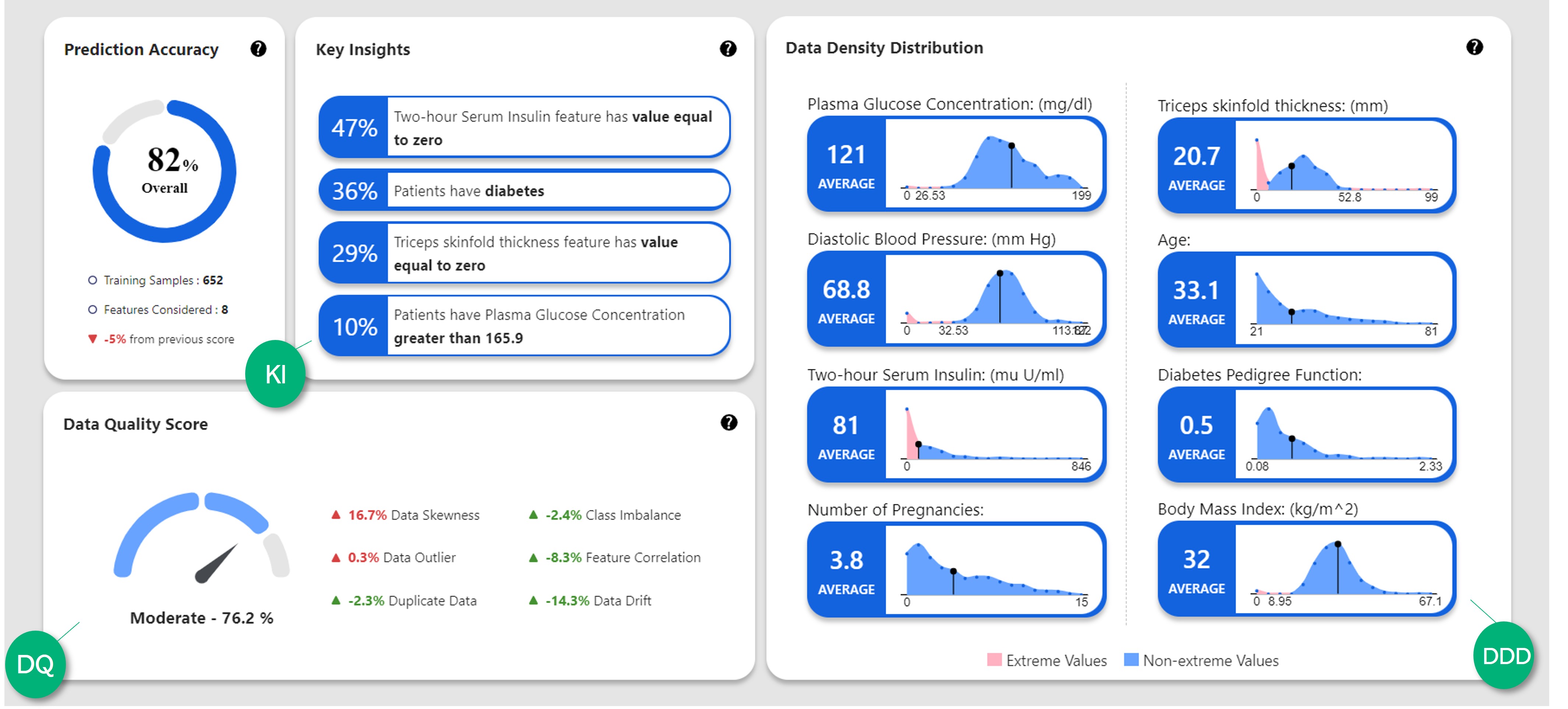}
\caption{Data-Centric Explanation dashboard design. Visual explanations are provided using: (KI) Key Insights, (DDD) Data Density Distribution and (DQ) Data Quality titles as marked in the figure.}
\Description[Data-Centric Explanation Dashboard]{Data-Centric Explanation dashboard design. Visual explanations are provided using: (KI) Key Insights, (DDD) Data Density Distribution and (DQ) Data Quality titles as marked in the figure.}
\label{fig:exmos_dce}
\end{subfigure}
\par\bigskip
\par\bigskip
\begin{subfigure}[b]{0.78\textwidth}
\centering
\includegraphics[width=1.0\linewidth]{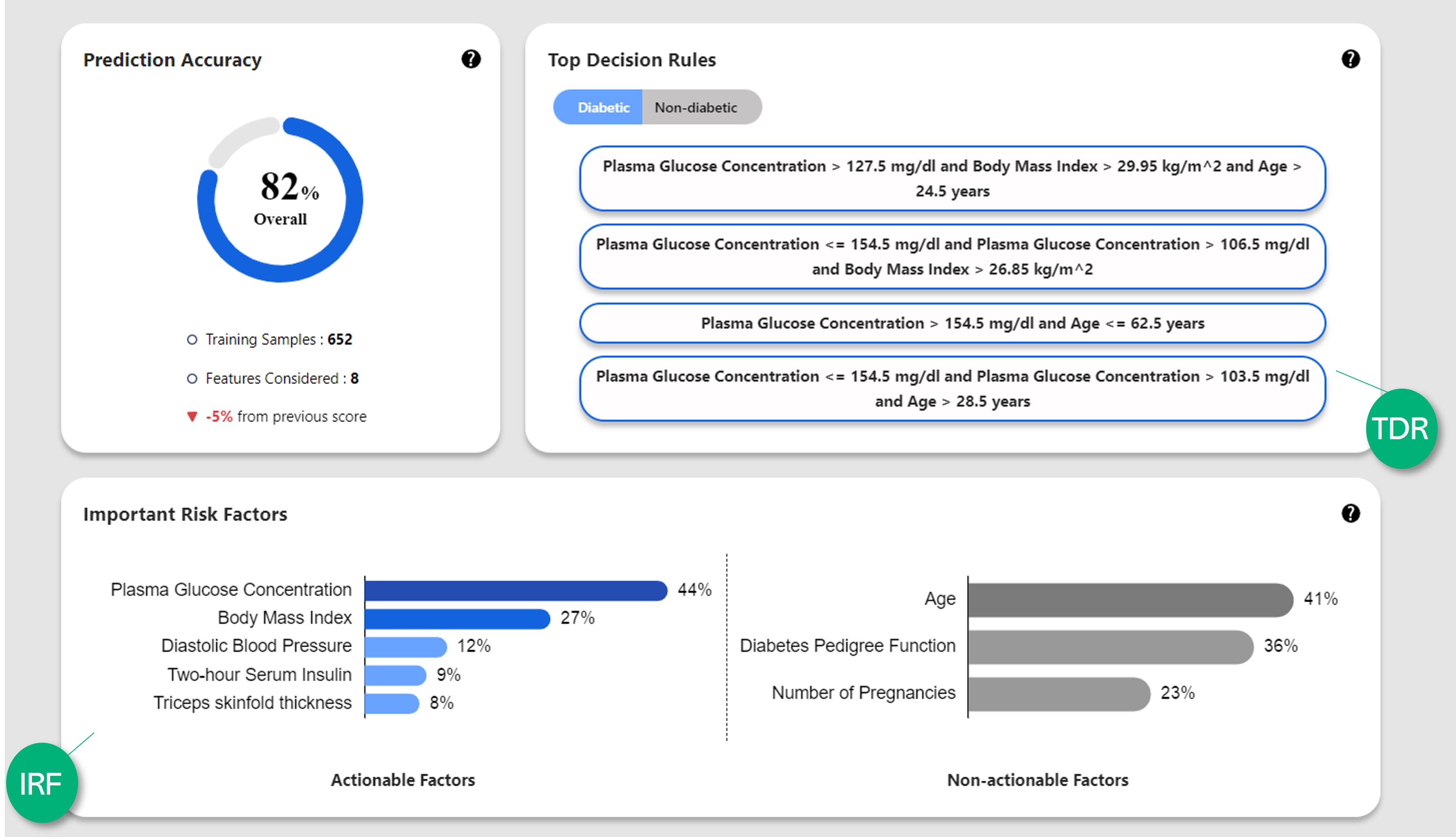}
\caption{Model-Centric Explanation dashboard design. Visual explanations are provided using: (TDR) Top Decision Rules, (IRF) Important Risk Factors titles as marked in the figure.}
\Description[Model-Centric Explanation Dashboard]{Model-Centric Explanation dashboard design. Visual explanations are provided using: (TDR) Top Decision Rules, (IRF) Important Risk Factors titles as marked in the figure.}
\label{fig:exmos_mce}
\end{subfigure}
\caption{Data-Centric and Model-Centric Explanation dashboards of our prototype }
\Description[Data-Centric and Model-Centric Explanation dashboards of our prototype]{Data-Centric and Model-Centric Explanation dashboards of our prototype}
\label{fig:exmos_dashboards}
\end{figure*}

\section{Design of Explanations and Model Steering Approaches}\label{sec_xai_design}

This section describes our generic designs of different explanation methods presented in a dashboard and of model steering approaches using manual and automated data configurations. We demonstrate an implementation of these designs for a healthcare-focused EXMOS system, as used in our user studies described in \Cref{sec_methods}.

\subsection{Explanation Dashboard}\label{sec_xai_dash}

Since this research aimed to compare diverse model-agnostic, global, model-centric and data-centric explanations, we designed the following three different versions of the explanation dashboard:
\par\bigskip

\textbf{(1) Data-Centric Explanation (DCE) version}: This dashboard version includes only global data-centric explanations, which offer insights into the overall patterns of the training data ~\cite{anik_data-centric_2021, Bhattacharya2023, BhattacharyaXAI2022}. \Cref{fig:exmos_dce} illustrates an implementation of this dashboard version for a healthcare-focused model steering system. Global data-centric explanations generally summarise the training data, highlight interesting findings, depict predictor variable distributions, and convey biases, inconsistencies, and data quality information.  The following visual components are designed to provide different types of global data-centric explanations:

\begin{itemize}
    \item \textbf{Key Insights (KI)}: This visual component aims to present insights about the training data generated using descriptive statistics. Primarily, this visual summarises the presence of biased and extreme (or anomalous) data values for each dataset variable using percentages. For our prototype, we displayed the top 4 insights on the dashboard tile, with additional insights provided as tooltips. The design of this explainability approach aligns with the ML transparency principle by Bove et al.\cite{bove_contextualization_2022} but is applied to global explanations.
    \item \textbf{Data Density Distribution (DDD)}: This visual component aims to present the distribution of value counts for each predictor variable, highlighting the average value, graphical data distribution, and detection of potential abnormalities (i.e. extreme values) from the training data. In our implementation, we included interactive tooltips to display the corresponding patient counts for each predictor variable. We utilised the design principles of data-centric explanations from Bhattacharya et al.~\cite{Bhattacharya2023} as its design rationale.
    \item \textbf{Data Quality (DQ)}: This visual component aims to depict the overall training dataset quality. The overall data quality can be estimated based on specific dataset issues such as outliers, redundant data, correlated features, class imbalance, data drift, data skewness and etc~\cite{lones2023avoid, ackerman2022detection, kazerouni2020active}. However, additional types of data issues may also be considered if present within the dataset. Each issue can be given equal weight for calculating the data quality score. In our implementation, the overall quality score is further abstracted into three levels: (1) good (\textit{if score > 80}), (2) moderate (\textit{if 50 $\leq$ score $\leq$ 80}), (3) poor (\textit{if score < 50}). Our design approach was aligned with Wang et al.~\cite{wang_designing_2019} and Bhattacharya et al.’s~\cite{Bhattacharya2023} approach for showing estimated uncertainty measures.
\end{itemize}

\textbf{(2) Model-Centric Explanation (MCE) version}: This version includes only global model-centric explanations, such as SHAP-based global feature importance~\cite{lundberg2017unified, adadi2018peeking} and surrogate explainer based top decision rules~\cite{Paul-Amaury2022, hepburn2021explainers}.  \Cref{fig:exmos_mce} illustrates an implementation of this dashboard version for a healthcare-focused EXMOS system. The following visual components are designed to provide different types of global model-centric explanations:
\begin{itemize}
    \item \textbf{Top Decision Rules (TDR)}: This visual component displays relevant decision conditions for predicting the different target classes present within the dataset. In our implementation, it presents the different rules for predicting patients as diabetic or non-diabetic. We applied the skope-rules Python module~\cite{skrules_python} to generate the top decision rules based on surrogate explainers. We utilised the causal attributions and inductive reasoning principles from Wang et al.~\cite{wang_designing_2019} for its design.
    \item \textbf{Important Risk Factors (IRF)}: This visual component displays global feature importance of the various predictor variables present within the dataset. In our implementation, the feature importance scores were generated using the Python SHAP module~\cite{shap_python}. The design principles of Bhattacharya et al.~\cite{Bhattacharya2023} can be followed to distinguish between actionable and non-actionable features while presenting the feature importances.
\end{itemize}

\begin{figure*}
\centering
\includegraphics[width=0.78\linewidth]{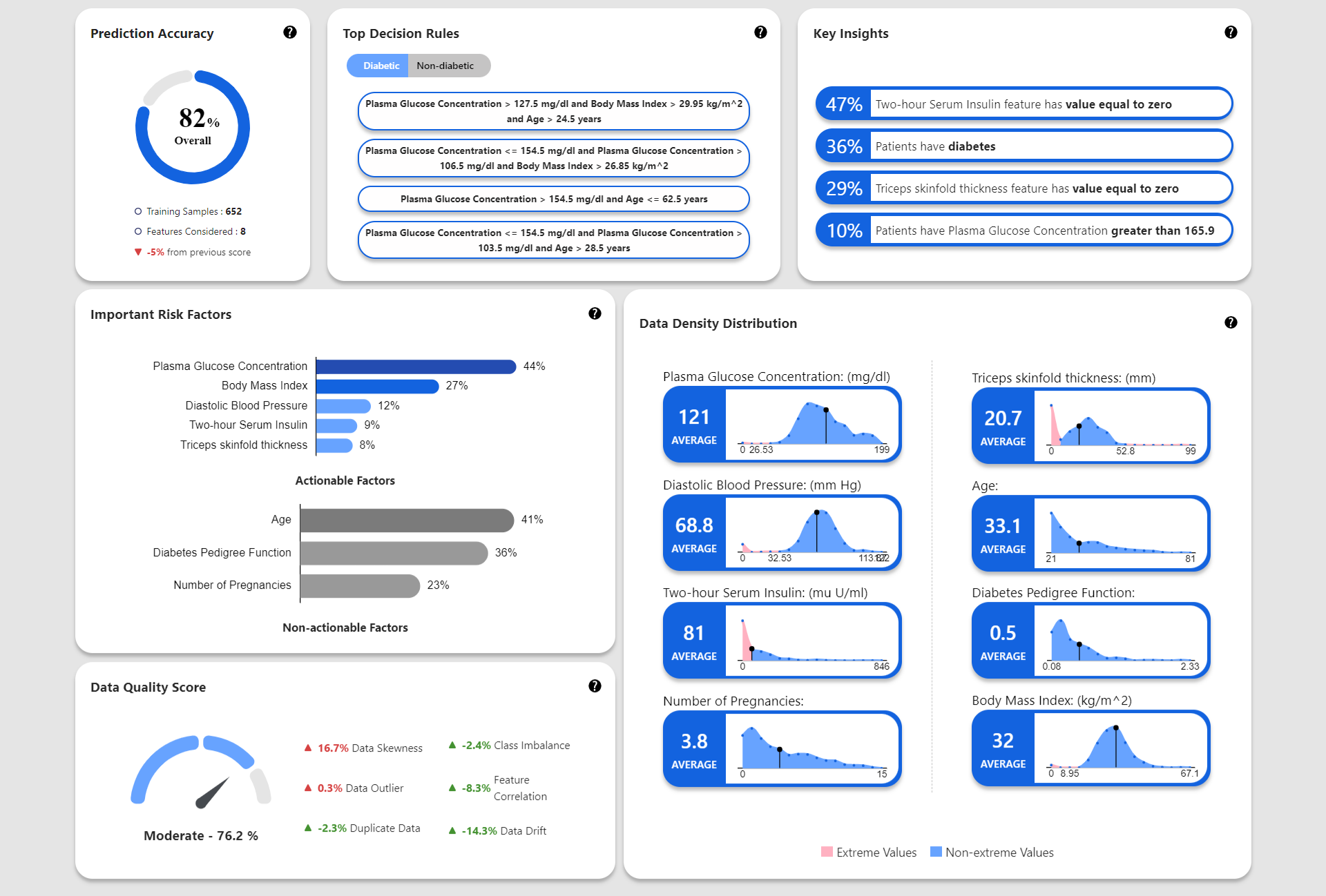}
\caption{Hybrid explanation dashboard: combining data-centric and model-centric explanations from \Cref{fig:exmos_dce} and \Cref{fig:exmos_mce}}
\Description[Hybrid Explanation Dashboard]{Hybrid explanation dashboard design: combining data-centric and model-centric explanations. This version included all five visual components: Key Insights (KI), Data Density Distribution (DDD), Top Decision Rules (TDR), Important Risk Factors (IRF), Data Quality (DQ)}
\label{fig:exmos_hyb}
\end{figure*}

\textbf{(3) Hybrid (HYB) version}: The design of the hybrid explanation dashboard basically combines all the visual explanations components providing global model-centric and data-centric explanations. \Cref{fig:exmos_hyb} illustrates our implementation of the hybrid explanation dashboard, which included all the different types of explanations from the DCE and MCE dashboards.

In our implementation, each version of the explanation dashboard presented the overall prediction accuracy to highlight the prediction model’s performance, the number of training samples, predictor variables (features) included in the training data and the percentage change from the previous version of the trained model. Furthermore, we included \faQuestionCircle \hspace{0.01cm} in each tile to assist our users by providing a description of each visual component. Our research involved comparing the DCE, MCE and HYB versions of the dashboard to address our research questions. However, these designs can be adapted to other use cases which involve explanatory model steering.

\begin{figure*}
\centering
\begin{subfigure}[b]{1.0\textwidth}
\centering
\includegraphics[width=0.8\linewidth]{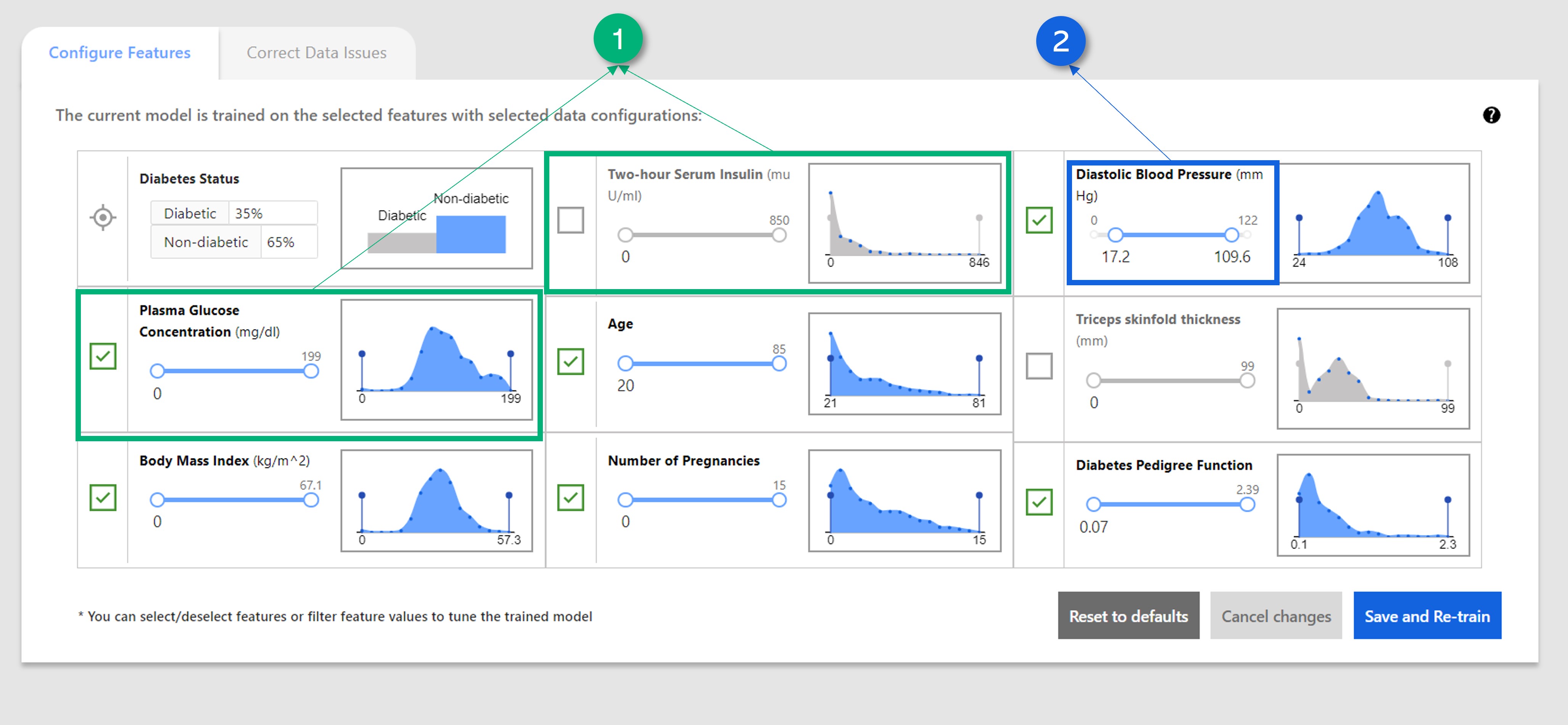}
\caption{Manual configuration screen from our prototype, which included (1) feature selection control to include or exclude predictor variables and (2) feature filtering control to set the upper and lower limits for the predictor variables. }
\Description[Manual configuration screen from our prototype]{Manual configuration screen from our prototype, which included (1) feature selection control to include or exclude predictor variables and (2) feature filtering control to set the upper and lower limits for the predictor variables.}
\label{fig:manual_config}
\end{subfigure}
\par\bigskip
\begin{subfigure}[b]{1.0\textwidth}
\centering
\includegraphics[width=0.8\linewidth]{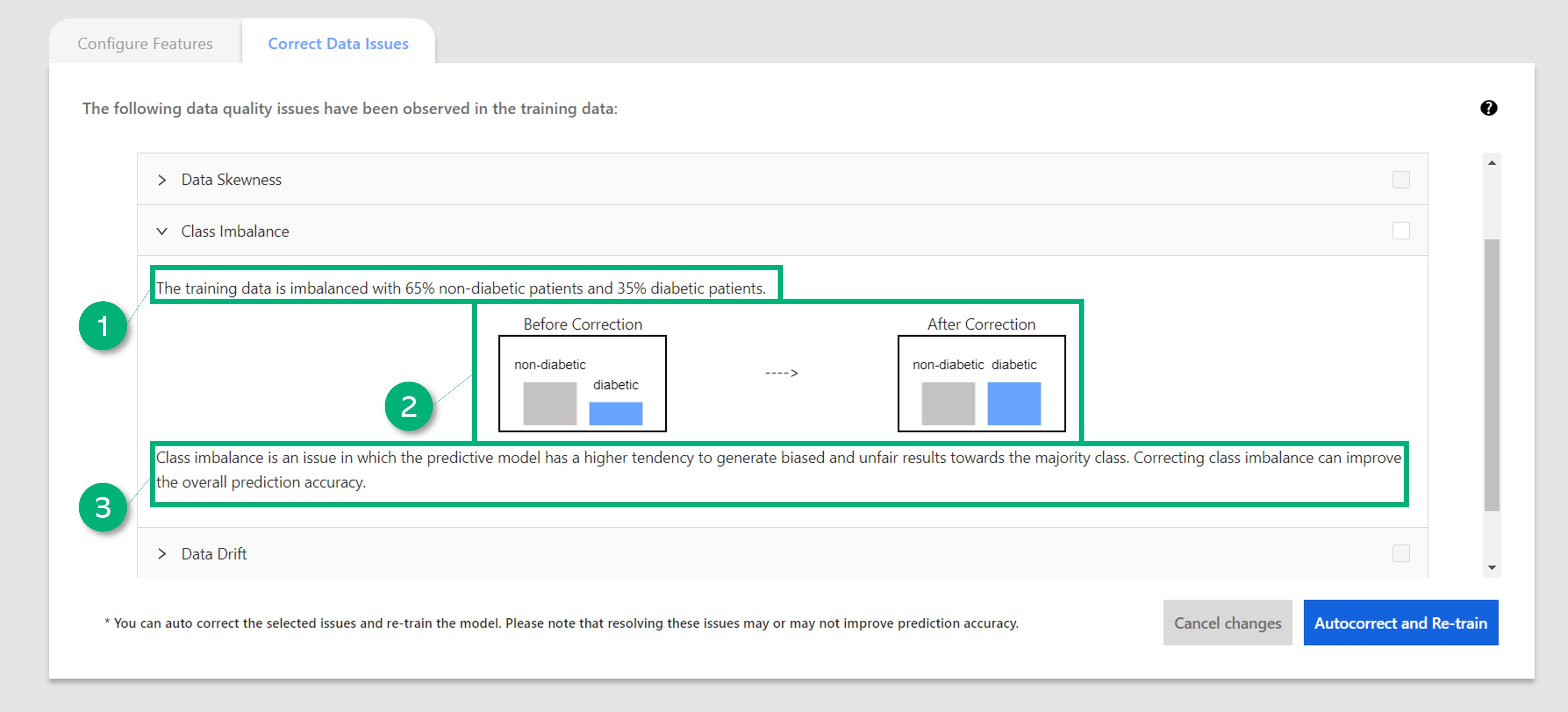}
\caption{Automated configuration screen: this screen explained the data issues through (1) displaying the quantified impact of these issues, (2) visualisations displaying before and after correction changes to the data or predictor variables and (3) description of the issue and how its correction can impact the model performance.}
\Description[Automated configuration screen from our prototype]{Automated configuration screen: this screen explained the data issues through (1) displaying the quantified impact of these issues, (2) visualisations displaying before and after correction changes to the data or predictor variables and (3) description of the issue and how its correction can impact the model performance.}
\label{fig:auto_config}
\end{subfigure}
\caption{Data configuration screens from our prototype. }
\Description[Data configuration screens from our prototype]{Data configuration screens from our prototype.}
\end{figure*}

\subsection{Data Configuration Mechanisms}\label{sec_data_config}
An essential part of our research involved studying how domain experts leverage their knowledge to retrain and improve the prediction model by configuring the training data. Thus, we designed the following configuration mechanisms which supports domain experts in model steering:
\begin{enumerate}
    \item \textbf{Manual configuration}: Using this mechanism, domain experts could select features and filter data by adjusting their upper and lower limits. The manual configurations provide more control to configure the training data, as domain experts could select or adjust the predictor variables using slider controls based on their requirements. For example, if domain experts, such as healthcare experts, think that diastolic blood pressure is not an important predictor variable for diabetes prediction, they can unselect the predictor variable to exclude it from the training data. Additionally, if they think that patients over 80 years old should be excluded from the training data, they can exclude them using the manual configuration. \Cref{fig:manual_config} illustrates the manual data-configuration screen implemented in our healthcare-focused EXMOS system.  We also included interactive visuals to display data distribution changes for each variable when configured in our implementation.  
    
    \item \textbf{Automated configuration}: The automated configuration mechanism prevents the manual workload of individually configuring the training data. This method aims to maximise the data quality by reducing various data issues, such as outliers, correlated features, skewed data, imbalanced categorical data and etc.~\cite{lones2023avoid, ackerman2022detection, kazerouni2020active}.  This method also involves application of automated correction algorithms to correct the data issues, such as the SMOTE technique [18 ] can be used to mitigate the class imbalance problem or removal of redundant data using automated algorithms. However, our design of this mechanism includes explanations of the data issues to increase the transparency of the automated correction algorithms. For each data issue, we recommend quantifying its impact on the overall data quality and displaying the estimated impact scores. We also recommend providing visualisations to demonstrate their impact on the data before and after the correction. Furthermore, we suggest adding simplified textual descriptions of each issue, justifying how their presence can affect the prediction model. These explanations of the data issues and the automated correction methods can enable users to identify potential issues that need correction and the model can be retrained after their automated correction.~\Cref{fig:auto_config} illustrates an implementation of our automated configuration design within our healthcare-focused EXMOS system.
\end{enumerate}
 
Furthermore, our designs allowed users to revert to the default data and model settings, discard unsaved changes and save and retrain the current changes, following guidelines established in \cite{kulesza_principles_2015}. After each configuration, the prediction model was regenerated with the configured data and all the explanations were re-calibrated. 

\section{Methods}
\label{sec_methods}
This section describes our study design, prototype implementation and the detailed methodology for two user studies.

\subsection{Study Design}

To compare the impact of the DCE, MCE and HYB explanation dashboards on trust, understanding and data-configuration mechanisms, we conducted a between-subject quantitative study (n=70) and another between-subject qualitative study (n=30) with healthcare experts as discussed in \Cref{sec_study1} and \Cref{sec_study2} respectively. The user studies were approved by the ethical committee of \anon{KU Leuven} with the number \anon{G-2021-4074}. The rationale for conducting both quantitative and qualitative user studies originated from the multifaceted objectives of our research. The primary focus of our quantitative study was to collect empirical evidence to determine which version of the explanation dashboard was more effective for improving prediction models, enhancing trust and understandability and influencing their selection of data configuration approach. However, through our qualitative study, we delved deeper to understand why and how different types of explanations support healthcare experts in model steering. Considering our research objectives, we only varied the version of the explanation dashboard in our between-subject user studies. Despite the automated configuration mechanism being a novel attempt to fine-tune prediction models by resolving common data issues, we were unsure how our participants would perceive it. Before extending the study's scope to include the diverse configuration approaches, we wanted to collect empirical evidence to support the usage of automated configurations for future research through our user studies. Therefore, different approaches to data configuration were not considered as study factors. Both manual and automated data configurations were supported in all three versions of the explanation dashboard.

\subsection{Prototype Implementation}\label{sec_proto_implementation}

We developed a high-fidelity web application prototype of an EXMOS system, in accordance with the designs outlined in \Cref{sec_xai_design}. The prototype provided global explanations to healthcare experts for the predictions generated by a diabetes prediction model using the DCE, MCE and HYB explanation dashboards illustrated in \Cref{fig:exmos_dashboards} and \Cref{fig:exmos_hyb}. The prototype enabled healthcare experts to make changes to the training data and retrain the model through manual and automated data configurations described in \Cref{sec_data_config}.  Next, we discuss the dataset and prediction model included in our prototype.

\emph{Dataset}: We leveraged the open-sourced Pima Indians Diabetes dataset\footnote{Source OpenML: \url{https://www.openml.org/search?type=data&sort=runs&id=37&status=active}} \cite{Smith1988-rv} for our prototype. The dataset comprises health records of 768 patients, 8 predictor variables and the target variable. It is primarily used for binary classification, i.e. classifying patients as diabetic or non-diabetic.  All the patients included in this dataset are Pima Indian women above 21 years old. This dataset was specifically chosen for our experiments due to its inherent data issues, such as an abnormally high number of zero values across numerous feature variables, outliers, an imbalanced distribution of target class, and skewed distributions observed in some of the features. We hypothesised that healthcare experts could understand the limitations of the data and build better prediction models by configuring the training data.

\emph{Prediction model}: Scikit-learn's~\cite{sklearn_api} implementation of the Random Forest algorithm was utilised for training a classifier on our diabetes prediction dataset, which resulted in a training accuracy of 84\% and a test accuracy of 80\%. The test accuracy was considered as the overall prediction accuracy in our prototype. During the model development phase, we experimented with several classification algorithms, such as logistic regression, support vector machines, k-nearest neighbours, XGBoost and deep neural networks. We also experimented with state-of-the-art AutoML tools such as Azure Automated ML~\cite{azureautoml}, PyCaret~\cite{pycaret}, Deep Neural Networks using AutoKeras~\cite{autokeras}.  However, the random forest model produced better and more generalised predictions as it had minimal overfitting and underfitting effects. Therefore, we selected this model for our final prototype. Our experimental results with different ML algorithms during the model development process are provided in the supplementary content. However, since our prototype included model-agnostic global explanations and steering approaches, the choice of the prediction model does not impact our design of the explanation dashboard or the data configuration mechanisms.

\subsection{User Study 1: Quantitative Study} \label{sec_study1}
\emph{Study setup}: We first conducted a between-subject quantitative study involving 70 healthcare experts to address our research questions. The study was conducted through the online survey platform Qualtrics~\cite{qualtric_online}. Previous studies that did not include a control condition have shown that participants respond positively only when explanations are presented~\cite{anik_data-centric_2021}. Thus, similar to other studies without a control condition~\cite{Dodge_2019, Kizilcec2016, kulesza_2013, anik_data-centric_2021}, we did not include a no-explanation condition in our study. Instead, our design focused on comparing the three different versions of the explanation dashboard.

\begin{table}[h]
\caption{Participant information for user study 1.}
\label{tab:participants_study1}
 \scalebox{.70}{
\begin{tabular}{ll}
\toprule
& \textbf{Participant Groups}                                                 
\\ \midrule
COHORT & \begin{tabular}[c]{@{}l@{}}
DCE: 25 \\
MCE: 27 \\
HYB: 18
\end{tabular}                                                                        
\\ \midrule
AGE GROUPS 
& \begin{tabular}[c]{@{}l@{}}
(18-29) years: 63\\ 
(30-39) years: 3\\
(40-49) years: 3\\ 
(50-59) years: 1 \\
%($mu$: 23, $sigma$: 6.7)%
\end{tabular}                                                                        
\\ \midrule
GENDER & \begin{tabular}[c]{@{}l@{}}
Female: 51 \\
Male: 17 \\
Non-binary: 1 \\
Not disclosed: 1
\end{tabular}                                                                        
\\ \midrule
HEALTHCARE EXPERIENCE  & \begin{tabular}[c]{@{}l@{}}
< 1 year: 13 \\
1-3 years: 3 \\
3-5 years: 33 \\
5-10 years: 17 \\
>10 years: 4
\end{tabular} \\ 
\bottomrule
\end{tabular}}
\end{table}

\emph{Participants}: We initially recruited 92 participants from a large network of volunteers from \anon{the Faculty of Health Science, University of Maribor}. We selected participants who were over 18 years old and had prior experience as healthcare assistants, paramedics, trainees, or registered nurses in the treatment and care of diabetes patients. Each participant was randomly assigned to one of the three versions (i.e. DCE, MCE or HYB). Prior to analysing the study results, we established two main exclusion criteria: (1) responses from participants who failed to answer all the survey questions were excluded, and (2) responses from participants who failed to complete the given tasks were excluded. After evaluating the study response, we excluded 22 responses considering these exclusion criteria. Thus, the results rely on data from a total of 70 participants, comprising 25, 27, and 18 participants for DCE, MCE and HYB, respectively. Additionally, to validate our sample size selection for each group, we conducted a power analysis based on standard guidelines~\cite{samplesizeguidelines}, resulting in a minimum sample size of 17 for each group to achieve a power of 0.85, maintaining an error rate below 0.05 and a medium effect size of 0.3. Moreover, our primary inclusion criterion was met as we were able to successfully include participants with varying levels of experience in healthcare, as presented in \Cref{tab:participants_study1}.

\emph{Study procedure}: Informed consent was obtained from participants along with detailed instructions on their roles, responsibilities, and rights for the study. Next, they were introduced to their allotted prototype version through tutorial videos describing the usage scenario, the purpose of the prediction model, the explanation dashboard and the data configuration mechanisms. Additionally, the participants were encouraged to explore the prototype independently after watching the tutorial videos to familiarise themselves with the system. After this, we collected their demographic information through survey questions. 

Next, the participants had to complete a model-steering task. In this task, participants were given 15 minutes to explore the system and perform training data configuration using the various configuration mechanisms, aiming to maximise the overall prediction accuracy from the default state. The participants were allowed to configure the training data multiple times. However, they were asked to switch to the configurations that gave them the maximum prediction accuracy before marking this task as complete. Post-completion of this task, we measured the perceived task workload using the NASA-TLX questionnaire~\cite{HART1988139, kulesza_principles_2015, kulesza_explanatory_2010}.

Additionally, our research aimed to study the impact of the different types of explanations on the understanding and trust of the users. We prepared a mental model questionnaire similar to Kulesza et al.~\cite{kulesza_principles_2015, kulesza_explanatory_2010} to measure the objective understanding~\cite{Cheng2019, bove_contextualization_2022} of users.
An answer was deemed correct if it matched the predefined expected response for each question. We also measured their subjective understanding using the perceived understandability questionnaire proposed by Hoffman et al.~\cite{hoffman2019metrics} and perceived trust using the Cahour-Forzy scale questionnaires~\cite{hoffman2019metrics} on a 7-point Likert scale. 
Additionally, we piloted our study to validate the working of the prototype and refine the vocabulary used in the study questionnaire.

\emph{Data collection and analysis}: The study collected the following types of quantitative data:
\begin{itemize}
    \item Updated prediction accuracy after the model steering task.
    \item Quantitative responses to the NASA-TLX workload assessment.
    \item Scores for the mental model questionnaire for evaluating the objective understanding.
    \item Perceived understandability responses on a 7-point Likert scale.
    \item Perceived trust responses on a 7-point Likert scale.
    \item System interaction data, such as mouse-click counts and hover time during interactions with the visual explanations and during manual and automated data configurations.
\end{itemize}

Since the recorded data violated the normality assumptions, we utilised non-parametric tests for statistical analysis. Specifically, we used the Kruskal-Wallis test~\cite{mccrum-gardner_which_2008} to assess overall significance, and for significant results, we conducted the Mann-Whitney U-test~\cite{mccrum-gardner_which_2008} with Bonferroni correction for pairwise group comparisons (DCE-MCE, MCE-HYB, and DCE-HYB). Descriptive statistics were also used for further analysis, and the results were visualised using comparative box-plots. Using the system interaction data, we computed average clicks per user (CPU) and average hover time per user (HTPU) to compare the three prototype versions. Similar to Verbert et al.~\cite{Verbert2016}, we also computed the \textit{effectiveness} and \textit{efficiency} of manual and automated configurations for each dashboard version to analyse how our participants used these different configuration approaches. \textit{Effectiveness} is measured by taking the ratio of successful attempts, where participants increase the model accuracy beyond the default value, to the total number of attempts made. \textit{Efficiency} is determined by calculating the ratio of the total hover time spent by participants to the total number of successful attempts made using each configuration type.

\begin{table}[h]
\caption{Participant information for user study 2.}
\label{tab:participants_study2}
 \scalebox{.70}{
\begin{tabular}{ll}
\toprule
& \textbf{Participant Groups}                                                 
\\ \midrule
COHORT & \begin{tabular}[c]{@{}l@{}}
DCE: 10 \\
MCE: 10 \\
HYB: 10
\end{tabular}                                                                        
\\ \midrule
AGE GROUPS 
& \begin{tabular}[c]{@{}l@{}}
(18-29) years: 28\\ 
(30-49) years: 2\\
\end{tabular}                                                                        
\\ \midrule
GENDER & \begin{tabular}[c]{@{}l@{}}
Female: 23 \\
Male: 7 \\
\end{tabular}                                                                        
\\ \midrule
HEALTHCARE EXPERIENCE  & \begin{tabular}[c]{@{}l@{}}
< 1 year: 1 \\
1-5 years: 18 \\
5-10 years: 11
\end{tabular} \\ 
\bottomrule
\end{tabular}}
\end{table}

\subsection{User Study 2: Qualitative Study}\label{sec_study2}

\emph{Study setup}: We conducted a between-subject qualitative study with 30 healthcare experts to gain deeper insights into their perceptions of utilising various explanation types during model steering. This study aimed to collect qualitative data to justify the quantitative results from the first study. 

\emph{Participants}: We recruited an additional 30 participants specialising in nursing and patient care from \anon{Faculty of Health Science, University of Maribor}. While the participants were not part of the first study, their selection criteria were the same as those of the first study. The recruited participants were randomly assigned to one of the three prototype versions, ensuring each version had 10 participants. \Cref{tab:participants_study2} presents additional demographic information about the participants.

\emph{Study procedure}: The study was conducted through face-to-face semi-structured individual interviews, which were recorded and transcribed for qualitative data analysis. The combined duration of all interviews amounted to approximately 1100 minutes, averaging around 35 minutes per interview.

After obtaining their informed consent, the participants were introduced to the prototype through a live demonstration of its functionalities. Then, each participant was given 5 minutes to independently explore the prototype, followed by semi-structured interviews. We referred to prior research work from Liao et al.’s XAI question bank~\cite{liao_questioning_2020}, Anik and Bunt~\cite{anik_data-centric_2021}, Cheng et al.~\cite{Cheng2019} and Kim et al.~\cite{KimCHI2023} to formulate the interview questionnaire. All questionnaires are included in the supplementary material. We also observed and recorded participant interactions with the prototype.

\emph{Data analysis}: The collected qualitative data was analysed using Braun and Clarke thematic analysis method~\cite{BraunClarkTA}. Using this method, we first reviewed the transcripts of the recorded interviews to identify an initial set of codes. Then, the identified codes were grouped into potential themes in several iterations. Upon reviewing the initial themes, we established a final set of themes to address our research questions. 

For participant anonymity, we referred to them as P(N), where N represents a specific participant number from 1 to 30. Only necessary grammatical corrections were made to the participant quotes when reporting the results.

\begin{table*}
\caption{Summary of statistical significance assessments. Statistically significant results are displayed in bold. CPU stands for clicks per user and HTPU stands for hover time per user.}
\label{tab:stats_summary}
 \scalebox{.8}{
\begin{tabular}{@{}c:c:ccc@{}}
\toprule
\textbf{Measures}                       & \textbf{Kruskal-Wallis test} & \multicolumn{3}{c}{\textbf{Mann-Whitney U-test}}                          \\ \midrule
                                        & \textbf{}                    & \textbf{DCE-MCE}            & \textbf{DCE-HYB}            & \textbf{MCE-HYB} \\       
\cmidrule(l){3-5}
\\
Post-task Prediction Accuracy    & \textit{\textbf{(H=10.89, p=0.004)}}  & \textit{(U=426.5, p=0.102)} & \textit{\textbf{(U=142.5, p=0.004)}} & \textit{\textbf{(U=109, p=0.001)}}  \\
Perceived Task Load (NASA-TLX) & \textit{\textbf{(H=12.52, p=0.0019)}} & \textit{(U=346, p=0.88)}    & \textit{\textbf{(U=109.5, p=0.004)}} & \textit{\textbf{(U=96, p=0.0007)}}  \\ 
\\
Objective Understanding        & \textit{(H=0.79, p=0.67)}    & \textit{(U=367.5, p=0.58)}  & \textit{(U=210.5, p=0.73)} & \textit{(U=206, p=0.39)} \\
Perceived Understandability       & \textit{(H=0.32, p=0.85)}   & \textit{(U=311, p=0.63)}    & \textit{(U=206.5, p=0.66)}  & \textit{(U=252.5, p=0.83)} \\
Perceived Trust               & \textit{(H=0.33, p=0.85)}   & \textit{(U=316.5, p=0.71)}  & \textit{(U=230, p=0.91)}    & \textit{(U=267.5, p=0.58)} \\ 
\\
Average CPU for the Explanation Dashboard & \textit{(H=0.63, p=0.73)} & \textit{(U=368.0, p=0.58)}    & \textit{(U=194.5, p=0.46)} & \textit{(U=231.0, p=0.79)}  \\  
Average CPU in Manual Configuration  & \textit{(H=0.25, p=0.88)}    & \textit{(U=358.5, p=0.71)}  & \textit{(U=227.0, p=0.72)} & \textit{(U=244.0, p=0.74)}  \\
Average CPU in Automated Configuration  & \textit{\textbf{(H=10.49, p=0.005)}}    &  {\textit{\textbf{(U=228.5, p=0.002)}}}  & \textit{(U=145.5, p=0.03) **} &\textit{(U=126.0, p=0.62)}  \\
\\
Average HTPU for the Explanation Dashboard & \textit{\textbf{(H=12.92, p=0.002)}} & \textit{(U=422.0, p=0.12)}    & \textit{\textbf{(U=313.5, p=0.003)}} & \textit{\textbf{(U=396.5, p=0.0004)}}  \\  
Average HTPU in Manual Configuration  & \textit{(H=2.48, p=0.29)}    & \textit{(U=347.5, p=0.86)}  & \textit{(U=265.5, p=0.32)} & \textit{(U=317.0, p=0.08)}  \\
Average HTPU in Automated Configuration  & \textit{(H=5.08, p=0.07)}   & \textit{(U=187.0, p=0.03) **}  & \textit{(U=85.5, p=0.31)} & \textit{(U=216.5, p=0.16)}  \\
\\
\bottomrule
\end{tabular}}
\begin{minipage}{12cm}
\vspace{0.1cm}
\small Note: (**) With Bonferonni correction, the significance level was adjusted to 0.0167 instead of 0.05 for Mann-Whitney U-test.
\vspace{0.1cm}
\end{minipage}
\end{table*}

\begin{figure*}[h]
  \centering
  \includegraphics[width=0.75\linewidth]{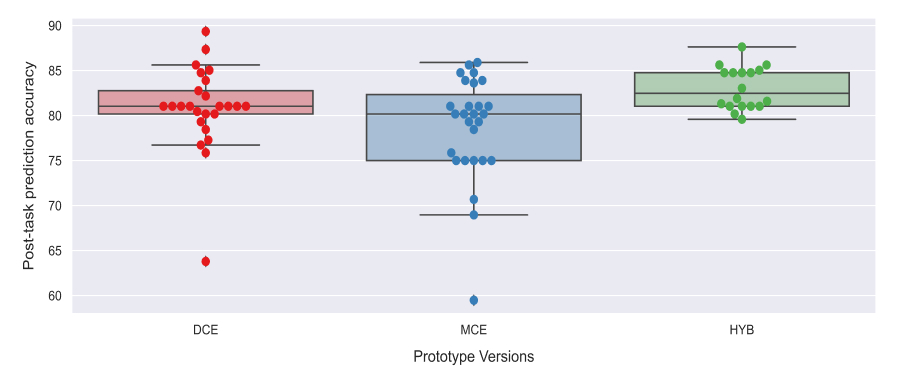}
  \caption{Box-plot showing the variation in the prediction accuracy scores obtained after the model steering task by participants of the three groups.}
  \Description[Box-plot showing Maximum Prediction Accuracy achieved by study 1 participants]{Box-plot showing post-task prediction accuracy obtained after the model steering task by participants. As shown in the figure, the HYB participants obtained much higher scores after the steering task than the DCE and MCE participants.}
  \label{fig:steering_task}
\end{figure*}

\section{Results}
This section presents the results of our user studies. \Cref{tab:stats_summary},~\ref{tab:dash_interactions} and~\ref{tab:add_measures_rq3} summarise the results from our quantitative study and \Cref{tab:themes_table} presents the themes generated from our qualitative study.

\textbf{RQ1. How do different types of global explanations affect healthcare experts’ ability to configure training data and enhance the prediction model’s performance, and why?} - Results from our first study indicate that HYB participants could significantly improve the prediction model performance compared to DCE and MCE versions. A Kruskal-Wallis test revealed a significant difference in post-task prediction accuracy scores across the three dashboard versions \textit{(H=10.89, p=0.004)}.~\Cref{fig:steering_task} illustrates a box-plot showing the post-task prediction accuracy obtained by participants from the three groups. Subsequent Mann-Whitney U-test with Bonferroni correction showed that the post-task prediction accuracy for HYB was significantly higher than MCE and DCE as the p-values between HYB-DCE and HYB-MCE were 0.004 and 0.001, respectively. Although no significant difference was found in scores between DCE and MCE users \textit{(U=426.5, p=0.102)}, 59.25\% of DCE participants were able to improve the prediction accuracy compared to only 40\% of MCE participants. However, as 88.8\% of HYB participants could improve the model accuracy,  combining data-centric and model-centric global explanations proved to be most impactful for model steering.

\begin{figure*}[h]
  \centering
  \includegraphics[width=0.9\linewidth]{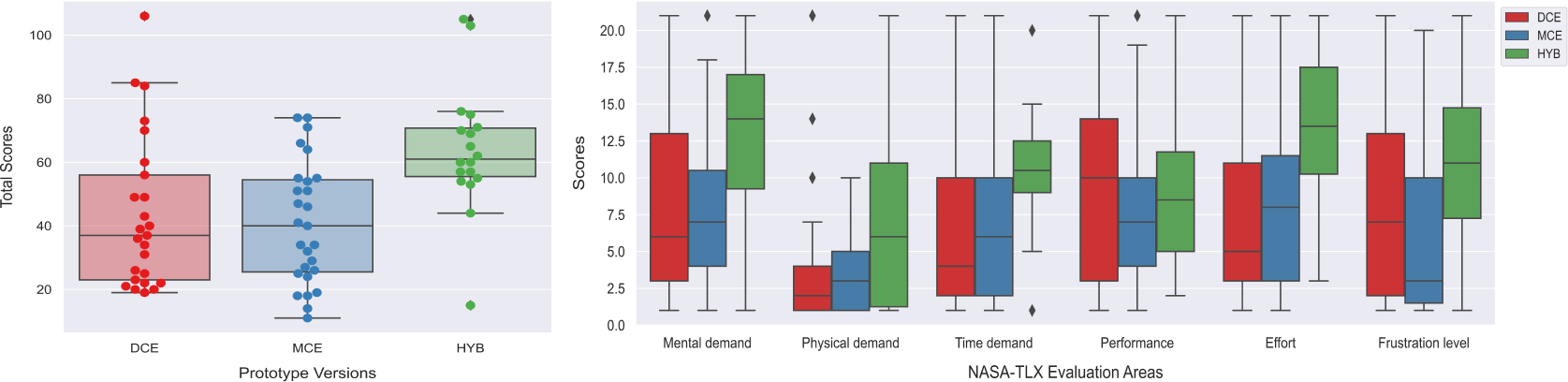}
  \caption{Box-plots showing variation of NASA-TLX scores between the three prototype versions and across the six evaluation areas of NASA-TLX such as mental demand, physical demand, time demand, performance, effort, frustration level. }
  \Description[Box-plot showing NASA-TLX scores]{Box-plots showing the variation of NASA-TLX scores between the three prototype versions and across the six evaluation areas of NASA-TLX such as mental demand, physical demand, time demand, performance, effort, frustration level. As shown in the figure, the HYB participants reported a much higher perceived task load across all the evaluation areas of NASA-TLX assessment than the DCE and MCE participants.}
  \label{fig:nasa_tlx}
\end{figure*}

Despite HYB participants achieving higher prediction accuracy, NASA-TLX task load assessments showed a significantly higher perceived task load for HYB participants than DCE and MCE \textit{(H=12.52, p=0.0019)}. Subsequent Mann-Whitney U-test with Bonferroni correction also showed a statistically significant difference between DCE and HYB versions \textit{(U=109.5, p=0.004)} and MCE and HYB versions \textit{(U=96, p=0.0007)} but not between DCE and MCE versions \textit{(U=346, p=0.88)}. The box-plots in~\Cref{fig:nasa_tlx} illustrate the overall variation of the perceived task load and the variation across each aspect of NASA-TLX. Additionally, from the user interaction data in~\Cref{tab:dash_interactions}, the average hover-time was significantly higher for the HYB version than the DCE \textit{(U=313.5, p=0.003)} or MCE \textit{(U=396.5, p=0.0004)} versions. These results indicate that merging these different types of explanations can initially overwhelm users.

\begin{table*}[h]
\caption{Summary of thematic analysis results. The themes are sorted in descending order of number of participants supporting them.}
\label{tab:themes_table}
 \scalebox{.8}{
\begingroup

\setlength{\tabcolsep}{10pt} % Default value: 6pt
\renewcommand{\arraystretch}{1.5} % Default value: 1
\begin{tabular}{|p{8cm}:p{7cm}|}
\hline
\textbf{Themes}                                                                                                            & \textbf{Supported by}                                                                                                                                                    \\ \hline
\textit{Transparency of data issues is important for better understandability and trust}                                            & \textbf{19 \faUserMd \hspace{0.2mm}   ($\sim$63\%)}: P4, P5, P9, P10, P11, P12, P14, P16, P17, P18, P19, P20, P21, P22,   P24, P25, P26, P27, P28 \\ \hline

\textit{Only global model-centric explanations are insufficient and non-actionable}                                                 & \textbf{16 \faUserMd \hspace{0.2mm}  ($\sim$53\%)}: P2, P3, P5, P8, P10, P12, P13,   P14, P16, P17, P20, P21, P24, P26, P29, P30                  \\ \hline

\textit{Global data-centric explanations improves system understandability post-configuration}                                & \textbf{14 \faUserMd \hspace{0.2mm}   ( $\sim$47\%)}: P1, P3, P4, P6, P7, P9, P10, P11, P12, P13, P14, P16, P19, P20                              \\ \hline

\textit{Manual configurations empower healthcare experts with flexibility, personalisation, and knowledge acquisition capabilities} & \textbf{14 \faUserMd \hspace{0.2mm}   ( $\sim$47\%)}:  P1,   P7, P9, P10, P11, P12, P13, P17, P18, P20, P21, P24, P26, P29                        \\ \hline

\textit{Information on data issues is crucial for medical researchers but less relevant for healthcare workers}                     & \textbf{13 \faUserMd \hspace{0.2mm}   ($\sim$43\%)}:  P5,   P6, P8, P10, P11, P12, P13, P16, P17, P18, P20, P21, P26                              \\ \hline

\textit{Data configurations positively impact the understandability and trust of the system}                                        & \textbf{12 \faUserMd \hspace{0.2mm}   ($\sim$40\%)}: P1, P3, P4, P7, P13, P16, P18, P19, P25, P27, P28, P30                                       \\ \hline

\textit{Auto-corrections are easier and faster  but less understandable than manual configurations}                                 & \textbf{12 \faUserMd \hspace{0.2mm}   ($\sim$40\%)}: P2, P5, P6, P11, P13, P16, P19, P20, P23, P24, P25, P27                                      \\ \hline

\textit{Local explanations can further enhance the usefulness and actionability of global explanations in healthcare}               & \textbf{11 \faUserMd \hspace{0.2mm}   ($\sim$37\%)}: P2, P3, P4, P5, P12, P13, P16, P17, P20, P21, P26                                            \\ \hline

\textit{Transparency about the data-collection process enhances trust}                             & \textbf{7 \faUserMd \hspace{0.2mm}   ($\sim$23\%)}: P3, P6, P9, P11, P15, P28, P30                                                                     \\ \hline

%\textit{Global model-centric explanations are also essential in healthcare}                                                         & \textbf{6 \faUserMd \hspace{0.2mm}   ($\sim$20\%)}: P2, P13, P17, P22, P23, P26                                                                   \\ \hline

\textit{The impact of sample size changes should be emphasised during data configurations}                                  & \textbf{6 \faUserMd \hspace{0.2mm}   ($\sim$20\%)}: P4, P6, P10, P14, P18, P25                                                                    \\ \hline

\textit{Collaboration and group configurations with peer approval mechanisms for effective data configurations}                     & \textbf{4 \faUserMd \hspace{0.2mm}   ($\sim$13\%)}: P8, P22, P25, P30                                                                             \\ \hline

\textit{Disclosing data issues is essential for decision-makers }                                                                   & \textbf{4 \faUserMd \hspace{0.2mm}   ($\sim$13\%)}: P9,   P11, P17, P28                                                                           \\ \hline

%\textit{Guided training is crucial for effective usage and adoption of healthcare XIL systems }                                     & \textbf{9 \faUserMd \hspace{0.2mm}   ($\sim$30\%)}: P2,   P4, P8, P10, P16, P18, P20, P27, P28                                                    \\ \hline%
%\textit{XIL systems should guide users to increase prediction accuracy along with data quality improvement}                         & \textbf{12 \faUserMd \hspace{0.2mm}   ($\sim$40\%)}: P3,   P4, P7,P10, P13, P16, P18, P19, P20, P23, P25, P29                                     \\ \hline%

%\textit{More data filters are needed for specific patient groups for better personalisation}                                        & \textbf{4 \faUserMd \hspace{0.2mm}   ($\sim$13\%)}: P8,   P13, P17, P25                                                                           \\ \hline

\textit{Importance of abstraction and gradual increase of detailed insights in data-centric explanations}                           & \textbf{3 \faUserMd \hspace{0.2mm}   ($\sim$10\%)}: P7, P11, P13                                                                                  \\ \hline

\textit{Maintaining a history of configurations and a roll-back mechanism for enhanced user experience}                             & \textbf{3 \faUserMd \hspace{0.2mm}   ($\sim$10\%)}: P10,   P19, P23                                                                               \\ \hline

%\textit{Data configurations in XIL systems is useful for collaboration between researchers, system developers, and healthcare experts}       & \textbf{17 \faUserMd \hspace{0.2mm}   ($\sim$56\%)}: P3,   P4, P5, P6, P8, P11, P13, P14, P16, P19, P20, P21, P23, P24, P26, P27, P28             \\ \hline
\end{tabular}
\endgroup}
\end{table*}

Our qualitative study delved deeper into the advantages and drawbacks of explanation types in model steering. Data-centric explanations proved valuable for healthcare experts in performing better configurations by providing a better understanding of the training data. Participants mentioned that all the visual components providing data-centric explanations (Key Insights, Data Density Distribution,
Data Quality) gave them a richer understanding of the training data. These explanations encouraged them to explore
the dataset more, eventually helping them to improve the prediction model through better configurations. For example,
P16 stated, \textit{``I could easily make changes in the data and observe changes in the dashboard ... it gives a better understanding of the system”}.

Moreover, we found global feature importance explanations to be insufficient and non-actionable to healthcare experts, who predominantly rely on their domain knowledge to conclude the importance of feature variables instead of algorithmic estimations of feature importance. For instance, P21 mentioned: \textit{``Glucose is the most important variable for diabetes, BMI generally increases with diabetes, but it is not very important. Some doctors who specialise in endocrine systems might consider Insulin as important, but nurses don’t consider it as important”}. Some participants found it hard to understand the change in prediction accuracy post-configurations using only model-centric
explanations: \textit{``The accuracy is higher and the explanations have changed, but I don’t understand why the accuracy is higher"} (P29). Additionally, some of our participants expressed the need for feature importance explanations to show the impact of specific features for elevating the risk of diabetes rather than the importance of the feature in the prediction process: \textit{``[From Important Risk Factors] these are just outcomes [importances], we would need the steps to reduce risk of diabetes ... it would be useful to know which risk factors we should target to reduce their risk of diabetes”} (P8).

Furthermore, our participants expressed a need for local explanations to enhance the usefulness and actionability of global explanations: \textit{``It will be more helpful to get individual patient information ... I will
make an exclusive healthcare routine for the specific patient.” (P2)}. Some participants also suggested that the increased perceived task load of hybrid explanations can be minimised by presenting an abstract high-level summary first and then a drill-down detailed view of the explanations so that it is not overwhelming for them at first sight: \textit{``It is better to present the data in a more simple manner first ... like a step-by-step approach in which they see very abstract information first, and then if someone is interested, they can go into the details of it”} (P11).

\colorlet{framecolor}{main}
\colorlet{shadecolor}{sub}
\setlength\FrameRule{0pt}
\begin{frshaded*}
\noindent\textbf{Key-takeaways}: Model-centric explanations alone are insufficient for facilitating domain experts in model steering. Conversely, data-centric explanations are more useful for domain experts in model steering due to the adequate elucidation of the training data. However, despite a higher perceived task load, the hybrid combination of data-centric and model-centric explanations proved most valuable for improving prediction model performance.
\end{frshaded*}

\textbf{RQ2. To what extent do different types of global explanations influence healthcare experts’ trust and understanding of the AI system?} - Findings from our first study did not reveal any significant differences in objective understanding~\textit{(H=0.79, p=0.67)}, perceived understanding~\textit{(H=0.32, p=0.85)} and perceived trust~\textit{(H=0.33, p=0.85)} across different prototype versions. However, as the scores did not significantly drop in the HYB version, we infer that the trust and understanding of healthcare experts in AI systems are not adversely affected by the increased perceived task load associated with merging data-centric and model-centric global explanations.

Unlike our quantitative study, our qualitative study revealed interesting insights about the perceived trust and understandability of users. Participants highlighted the importance of data-centric explanations in understanding system changes post-configuration by visualising predictor variable distributions and data quality changes. They mentioned how highlighting data issues further justified the data quality score, enhancing system transparency: \textit{``Showing the data quality improves the transparency of the system. If it is not higher, we can use this system with the researchers [or developers] to make the data quality higher”} (P20). Moreover, the users could improve the training data by removing abnormal data values and selecting relevant feature variables via configuration mechanisms: \textit{``The extreme values make us wonder why the zero values are higher and we want to understand more. It indicates that maybe it’s not the patients, but instead, there’s something wrong in the system.”} (P13). Consequently, they had higher confidence and trust in relying on the predictions. Also, some of our participants suggested that including information on the data collection process can boost the transparency and trust of the system: \textit{``Better to explain more about data source, how it was collected to explain the chances of occurrences of such issues”} (P30).

Additionally, as observed in our second study, training data visualisation on the configuration page facilitated user exploration and experimentation, thus improving system understanding and trustworthiness: \textit{``The data configurations allowed me to explore the system and have more control over it. It makes it more understandable and trustworthy”} (P9), \textit{``I trust the system better now as there are no extreme values after I have removed them [after data configuration]”} (P27).

\colorlet{framecolor}{main}
\colorlet{shadecolor}{sub}
\setlength\FrameRule{0pt}
\begin{frshaded*}
\noindent\textbf{Key-takeaways}: Despite the increased perceived task load, trust and understanding of domain experts are not adversely affected by the hybrid combination of global explanations during model steering. Particularly, global data-centric explanations helped domain experts to understand post-configuration system changes. Providing explanations about the data quality and disclosing the data issues allowed our users to have higher trust and confidence in the predictions. Moreover, allowing domain experts to configure the training data enabled them to explore the system more and have a better understanding of it. Additionally, disclosing the data collection process can further enhance their trust in the system.
\end{frshaded*}

\begin{table*}[h]
\caption{User interaction data for different explanation types and configuration mechanisms. Average Clicks Per User (CPU) is measured in the number of mouse clicks, and average Hover Time Per User (HTPU) is measured in seconds.}
\label{tab:dash_interactions}
 \scalebox{.75}{
\begin{tabular}{@{}c:cc:cc:cc@{}}
\toprule
                                             & \multicolumn{2}{c}{\textbf{DCE}}            & \multicolumn{2}{c}{ \textbf{MCE}}            & \multicolumn{2}{c}{\textbf{HYB}}            \\ \midrule
\\
\textbf{Average Clicks Per User (CPU)}      & \multicolumn{2}{c}{39.24}                                 & \multicolumn{2}{c}{33.96}                                 & \multicolumn{2}{c}{41.22}                                 \\ 
\textbf{Average Hover Time Per User (HTPU)} & \multicolumn{2}{c}{283.56 }                        & \multicolumn{2}{c}{156 }                           & \multicolumn{2}{c}{534.22 } 
\\\\ \hdashline
                                             & \textit{Avg. CPU } & \textit{Avg. HTPU } & \textit{Avg. CPU } & \textit{Avg. HTPU } & \textit{Avg. CPU } & \textit{Avg. HTPU} \\ \hdashline
\\
\textbf{Key Insights}                                 & 4.6                        & 24.04                        & -                          & -                            & 1.9                        & 25.55                        \\
\textbf{Data Density Distribution }                   & 5.5                        & 25.64                        & -                          & -                            & 1.5                        & 37.11                        \\
\textbf{Data Quality}                          & 2.2                        & 14.36                        & -                          & -                            & 2.0                        & 16.47                        \\
\textbf{Important Risk Factors }                      & -                          & -                            & 4.0                        & 14.11                        & 2.0                        & 27.11                        \\
\textbf{Top Decision Rules }                          & -                          & -                            & 5.78                       & 19.76                        & 3.7                        & 28.27                        \\ 
\\
\hdashline
\\
\textbf{Manual Configuration}    & 26                    & 635                    & 19                    & 680                    & 17                  & 527                    \\
\textbf{Automated Configuration} & 4                     & 609                    & 2                     & 381                    & 2.5                   & 488 
\\
\\
\bottomrule
\end{tabular}}
\end{table*}

\begin{table*}[h]
\caption{Table presenting \textit{effectiveness} and \textit{efficiency} of manual and automated data configurations for different explanation types.}
\label{tab:add_measures_rq3}
 \scalebox{.9}{
\begin{tabular}{@{}c|cc:cc:cc@{}}
\toprule
                                             & \multicolumn{2}{c}{\textbf{DCE}}            & \multicolumn{2}{c}{ \textbf{MCE}}            & \multicolumn{2}{c}{\textbf{HYB}}            \\ \midrule

                                             & \textit{Effectiveness } & \textit{Efficiency } & \textit{Effectiveness} & \textit{Efficiency} & \textit{Effectiveness } & \textit{Efficiency } \\ 
                                             \cmidrule(l){2-7}
                                             \\
\textbf{Manual Configuration}    & 0.64                    & 17.72                    & 0.46                    & 42.23                    & \textbf{0.75}                  & \textbf{15.28}                    \\
\textbf{Automated Configuration} & 0.43                     & 61.58                    & 0.34                     & 62.25                    & 0.59                   & 33.08 
\\
\\
\bottomrule
\end{tabular}}
\end{table*}

\textbf{RQ3. How do different types of global explanations impact the choice of steering models through training data configuration?} - As presented in~\Cref{tab:dash_interactions}, analysis of user interaction data from our quantitative study revealed that HYB users were faster to perform manual configurations than DCE and MCE users, with fewer interactions while configuring the training data. Initially overwhelmed by diverse explanations on the dashboard, HYB users spent more time exploring the interface. However, they were able to perform faster and better model steering through manual configurations. As shown in \Cref{tab:add_measures_rq3}, the HYB users demonstrated the highest \textit{effectiveness} and the best \textit{efficiency} for both manual and automated configurations than the DCE and MCE users. The MCE users obtained the lowest \textit{effectiveness} and the worst \textit{efficiency} for both manual and automated configurations. Despite additional effort needed in manual configurations, as presented in \Cref{tab:add_measures_rq3}, participants from all three versions were more \textit{effective} and \textit{efficient} when configuring the data manually rather than using automated configurations. These results highlight the importance of providing domain experts with more control over the prediction system.

While MCE users invested more time in manual configurations, they were not as successful as the HYB and DCE users in improving the prediction model. This result highlights the insufficiency of global model-centric explanations towards guiding users in model improvement. Furthermore, due to the absence of data quality information on the MCE version, the MCE users invested less time in exploring the automated configurations in which the issues in the training data were resolved automatically. In contrast, DCE and HYB versions provided a better understanding of the automated configurations by delineating the data quality and explaining the data issues. Consequently, DCE and HYB users were more \textit{effective} and \textit{efficient} in automated configurations than the MCE users.

Findings from our qualitative study helped us understand why our participants preferred more control over the training data through the manual configurations. They mentioned that the manual configurations provided more flexibility in selecting relevant patient groups and allowed them to experiment with the training data. For instance, P21 mentioned, ``\textit{the manual control is more useful than the automated one, as the automatic changes might be good for more general patients, while for a specific group of patients, the manual configurations give more control to have reliable results}". The manual configurations also helped them to learn the impact of certain health measures on the risk of diabetes for specific patient groups: ``\textit{It’s also a good tool for learning if medical students are not so well aware of diabetes, they can play around and learn how each factor can affect the diabetes predictions}" (P21). Furthermore, P29 mentioned that observing changes after the manual configuration helped to validate their own knowledge of how certain parameters can impact the risk of the disease: ``\textit{As a nurse, you might have the knowledge but some approval or validation of the knowledge makes you feel more confident. And this system [with manual configurations] helps you to apply this knowledge and see the changes in predictions.}”

However, the majority of the MCE participants (60\%) from our qualitative study had expressed scepticism about the manual configurations as they feared that incorrect data configurations could lead to poor predictions. Thus, instead of relying on individual feedback, they suggested that healthcare workers can feel more confident in performing data configuration as a group: ``\textit{It would be better to introduce this to a group of nurses [healthcare experts], then they would feel more confident to decide certain parameters, set the limits and turn on and off the risk factors as a group}” (P8). As a solution to this problem, P25 provided an interesting suggestion of having a peer review and approval system for data configurations: ``\textit{The system should allow individual users to suggest changes. Then the responsible group of nurses and doctors can see these suggested changes and perform these changes to see the benefits, otherwise decline the suggested changes.}”

\colorlet{framecolor}{main}
\colorlet{shadecolor}{sub}
\setlength\FrameRule{0pt}
\begin{frshaded*}
\noindent\textbf{Key-takeaways}: The hybrid combination of global explanations demonstrated the highest \textit{effectiveness} and \textit{efficiency} in model steering over the other versions. Empowering domain experts with greater control during model steering proves crucial, as participants across all three versions demonstrated more \textit{effective} and \textit{efficient} steering through manual configurations compared to the automated ones. Yet, concerns about potential user-induced errors during manual configurations express a need for group configurations and peer approval mechanisms.
\end{frshaded*}

\section{Discussion}

\subsection{Combining Different Types of Explanations for Effective Data Configurations} 

While our results favour global data-centric explanations over their model-centric counterparts, we advocate a hybrid approach, acknowledging the relevance of global model-centric explanations. Considering the \textit{No Free Lunch} theorem in ML~\cite{goldblum2023free, Wolpert1997}, we conjecture its applicability to XAI, where certain methods can elucidate only specific dimensions of explainability. Furthermore, as our HYB participants were the most successful in improving the prediction model, merging both types of explanations can facilitate users to obtain the most optimal prediction models for varied use cases. 

We also propose including interactive visualisations that summarise the training data, highlight interesting patterns in the data, display the density distribution of the feature variables, delineate the data quality and describe the data collection process for the data-centric explanations. Regarding global model-centric explanations, feature importance explanations should elucidate the impact of specific features for elevating the risk of the medical condition (e.g., diabetes) instead of simply showing the importance percentage of the feature in the prediction process. We draw inspiration from Wang et al.~\cite{wang_designing_2019} and Bhattacharya et al.’s~\cite{Bhattacharya2023} work for designing these explanations. 

Moreover, our qualitative study participants expressed a need for local explanations. A fusion of global and local explanations can be most useful for domain experts as local explanations provide a better contextualisation of global explanations~\cite{Wang_deepseer2023}, which we believe is essential for successful model steering. For example, local data-centric explanations can bolster the validity of the top decision rules by allowing users to access specific records that fulfil the rule. Similarly, spotlighting abnormal data points in a specific record along with the entire feature variable can help users distinguish between the corrupted records and noisy feature variables that should be excluded from the training data.

\subsection{Importance of Manual and Automated Data Configurations in EXMOS Systems} 

Although the goal of this research was to investigate the impact of different global explanations during model steering, our results revealed the importance of the different data configuration mechanisms included in our prototype. We found that these feedback mechanisms encouraged users to explore the system better and, consequently, develop a better mental model of the explanations and the prediction model. Additionally, data configuration mechanisms can improve collaboration between medical researchers, system developers and healthcare experts, who otherwise work in silos, as they can effectively collaborate to understand the data and the prediction models.
Furthermore, we recommend EXMOS systems to include both manual and automated configurations as different users prefer different levels of control during data configurations. However, future research should study the impact of these different data configuration mechanisms in isolation to better assess their impact on trust and understandability.

\subsection{Design Guidelines for Explanations and Data Configuration Mechanisms for Model Steering}
Based on the observations, results and participant feedback from our user studies, we propose the following guidelines for the design of explanations and data configuration mechanisms for model steering by domain experts:
\begin{itemize}
    \item \textit{Combining global data-centric and model-centric explanations}: The results of our user studies have highlighted the importance of combining global data-centric and model-centric explanations. Therefore, we recommend combining these different explanation types to empower healthcare experts for effective model steering.
    \item \textit{Including local explanations to enhance the usefulness, understandability and actionability of global explanations}: Considering the feedback of our participants, we propose combining local explanations with global explanations to enhance their usefulness, understandability and actionability. We posit that incorporating different types of local explanations, including counterfactual explanations, what-if explanations, local data-centric and model-centric explanations, as outlined in Bhattacharya et al.’s~\cite{Bhattacharya2023} work, can elevate the actionability of global explanations.
    \item \textit{Importance of abstraction when providing visual explanations}: Since an excessive number of visualisations in a single view can be confusing and overwhelming, we suggest showing only high-level summary information in the initial view. It can include the sample size of the training data, feature variable descriptions, prediction variable information, and overall model performance. Elaborated global and local explanations should be reserved for subsequent views.  Any technical details, such as the descriptions of the diverse data issues, should be presented in cascaded drill-down views. Furthermore, seamless navigation from global explanations to local explanations should be ensured. The layer of abstraction can prevent visual information overload and encourage more effective user exploration of the interface.
    \item \textit{Detailed information on data issues is crucial for decision-makers and researchers, but it should not be in the main explanation dashboard}: Our participant feedback underscores the significance of comprehensive data quality information for decision-makers such as doctors, lead nurses and researchers involved in medical experimentation and validation. Nevertheless, this information is irrelevant for non-decision makers like health workers operating in clinical settings. Thus, we recommend presenting this information in secondary or tertiary drill-down views of the explanation dashboard instead of the primary view.
    \item \textit{Importance of disclosing the data collection process}: We advocate the importance of disclosing the data collection process to elucidate the occurrences of abnormal data values and noisy feature variables. This information can be shown in the initial high-level summary view. 
    \item \textit{Allowing easy roll-back to any previous version during data configuration}: We suggest maintaining the data configuration history and implementing a roll-back mechanism to any previous configuration settings to promote higher adoption of EXMOS systems in high-stake domains, such as healthcare. %In light of participant feedback, we emphasise the significance of offering both manual and automated data configurations, recognising diverse user preferences for control levels. 
    Additionally, users should be able to override automated configurations, revert to default settings and undo changes seamlessly. %Automated configurations should also automatically remove abnormal data values, corrupt records and noisy feature variables. 
    However, the system should issue warnings if data configurations lead to a substantial reduction in training samples, as it can cause the prediction model to overfit or underfit the data. 
    \item \textit{Importance of peer approval in model steering through manual configuration}: Along with an easy roll-back option, we recommend providing a peer approval functionality, such that proposed changes in the training data through manual configurations can be reviewed by a panel of domain experts.  With this approach, individual users can propose changes,  while a panel of approvers comprising of decision-makers such as lead doctors, lead nurses or medical researchers can review, accept or even decline the proposed changes. Such a group consensus process can safeguard against the removal of important training data, preventing adverse effects on prediction models during model steering through peer approval.

\end{itemize}

\subsection{Limitations}
The following are some limitations of this work: 

(1) \textit{Institute-Centric Participant Recruitment}: Although participants from the first study did not partake in the second one, the recruitment was limited to the same institute for both studies. This localised recruitment strategy could introduce potential biases that we need to address in future research.

(2) \textit{Quantitative Study Sample Size}: Although we validated our participant sample size for our first study using standard guidelines~\cite{samplesizeguidelines}, a broader study could reveal deeper insights into different explanation methods and configuration mechanisms, thereby elevating the statistical power for the obtained results. Considering the limited availability of healthcare experts, we faced limitations in expanding our participant pool for the quantitative study.

(3) \textit{Limitations Due to Participant Age Range}: The majority of the participants in both of our user studies were between 18-29 years. Despite being successful in including participants with varied healthcare experience, the recruitment of older individuals for our studies was limited. Consequently, the insights and feedback from older age groups are not comprehensively addressed in our work.

(4) \textit{Unexplored Variation in Data Configuration Mechanisms}: In our user studies, we did not consider the different data configuration mechanisms as another variable factor along with the different types of explanations as this research focused on only exploring the impact of different explanations. However, we acknowledge that it is important to study the impact of these different data configurations on trust and understanding of explanations and prediction models as it could reveal more insights about their benefits and disadvantages.

\subsection{Future Work}
In our future work, we plan to address these limitations while pursuing new avenues for the improvement of EXMOS systems. We will conduct a randomised control study to gauge the impact of the data configurations on trust and understanding of the explanations and model improvement. We will also investigate the joint influence of global and local explanations in EXMOS systems. Our qualitative study highlighted the importance of combining global and local explanations, but we want to collect more quantitative evidence to analyse the coexistence of global and local explanations in EXMOS systems. For our future work, we also aspire to investigate the influence of conversation-based explanations~\cite{lakkaraju2022rethinking, Slack2023}. and data configuration mechanisms during model steering. We hypothesise that conversation-based approaches will have a lower perceived task load for improving prediction models than our current manual configurations approach.

\section{Conclusion}
Our work introduces an EXMOS system that utilises diverse explanation types to elucidate a diabetes prediction ML model. This system empowered healthcare experts to fine-tune the model via both manual and automated data configurations, leveraging domain knowledge. We delved into the influence of data-centric and model-centric global explanations during model steering, targeting improved prediction models and measuring their impact on trust, understanding, and data configuration approach. Findings from our user studies with healthcare experts indicate that global model-centric explanations are insufficient and non-actionable. Data-centric global explanations outperformed their model-centric counterparts, particularly in understanding post-configuration system changes. Nonetheless, combining data-centric and model-centric global explanations proved more \textit{effective} and \textit{efficient}. Based on our analysis, we share our design guidelines for explanations and data configuration mechanisms for explanatory model steering. Our work emphasises the importance of multifaceted explanations and domain-expert driven data configurations for model steering. 

%% identification of the section in the article metadata
\begin{acks}
We would like to thank Ivania Donoso-Guzmán, Maxwell Szymanski, Robin De Croon for providing helpful comments that improved this article. We extend our thanks to Robert Nimmo from the University of Glasgow, UK for helping us with the study design for our first user study. We also thank our colleagues from the Faculty of Health Science, University of Maribor, Slovenia, for making the necessary arrangements for our second user study. This research was supported by Research Foundation–Flanders (FWO, grants G0A4923N and G067721N) and KU Leuven Internal Funds (grant C14/21/072).
\end{acks}

\bibliographystyle{ACM-Reference-Format}
\bibliography{references}



\section*{Supplementary material}

# APPENDIX

## A. Technical Implementation

**Prototype Code and Demo Video:**

The source code for our React.js based front-end web application, FastAPI Python based backend application and deployment-ready docker configurations are available on GitHub: https://github.com/adib0073/EXMOS. Please update necessary constant values, such as, the Mongo DB connection string, collection names and application URL in *EXMOS > app-api > app > constants.py* and *EXMOS > app-ui > imports > ui > Constants.jsx* to successfully launch the applications.

**Experimental Results of Machine Learning Algorithms:**

We explored the following algorithms to build our machine-learning prediction model:
- Logistic Regression
- Support Vector Machines (SVM)
- K-Nearest Neighbours (kNN)
- Deep Neural Networks (DNN)
- Random Forest
- XGBoost

We selected accuracy as the metric for evaluating the models.

| Algorithm | Training accuracy | Test accuracy |
|---|---|---|
| Logistic Regression | 76% | 79% |
| SVM | 79% | 78% |
| KNN | 80% | 77% |
| **Random Forest \*\*** | **84%** | **80%** |
| XGBoost | 100% | 73% |
| DNN | 80% | 77% |

Table A1: Training accuracy and test accuracy for different algorithms after necessary hyper-parameter tuning for the ML model for our XIL prototype system

We also experimented with state-of-the-art AutoML tools such as Azure Automated ML [55], PyCaret [62], Deep Neural Networks using AutoKeras [39]. The results of the best models from these AutoML tools are as follows:

| AutoML Tool | Algorithm | Training accuracy | Test accuracy |
|---|---|---|---|
| Azure Automated ML | Random Forest | 83% | 78% |
| PyCaret AutoML | Logistic Regression | 81% | 77% |
| AutoKeras | Structured Data Classifier (AutoKeras) | 79% | 73% |

**\*\* As presented in the table, the Random Forest model had the least overfitting and underfitting effects. It was more generalised. So, we selected the Random Forest models for our prototype. However, since our prototype included model-agnostic global explanations and steering approaches, the choice of the prediction model does not impact our design of the explanation dashboard or the data configuration mechanisms.**

## High-Level Solution Architecture:

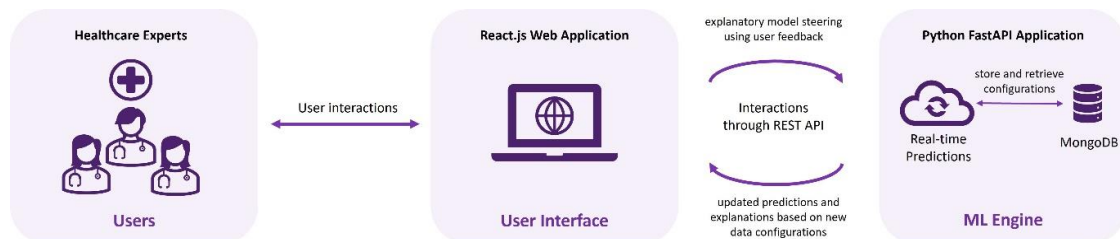

Our prototype XIL system allowed healthcare experts to interact with a React.js web application that provided explanations and allowed users to perform data configurations. The web application interacted with a Python FastAPI application connected to a MongoDB database through REST API calls. Separating the front-end user interface from the backend ML engine helped us achieve real-time predictions and model retraining and updated explanation generations without compromising the system performance.

# B. User Study 1: Quantitative Study Protocols and Questions

## Study Introduction
- Informed Consent
- Context Setting
- Prototype demonstration through tutorial videos
- Self-exploration of prototype

## Demographic Questions
1. Age
2. Gender
3. Occupation
4. Experience in healthcare
5. Field of specialisation

## Model Steering Task

The participants were given a model steering task in which they could perform any configurations on the training dataset and re-train the model. Their goal was to maximise the prediction accuracy after configuring the training data. They were given 15 mins time to complete this task.

They could configure the ML model by the following approaches:

1. They could select/unselect risk factors and retrain the model with only relevant risk factors which do not have any data issues (feature selection: manual configuration).

2. They could also filter and select the range of data values for each risk factor to remove any abnormal value from the training data and retrain the system (feature filtering: manual configuration).

3. They could also try auto-correcting the data issues and retrain the model to observe if the prediction accuracy improved (auto-correction).

## NASA-TLX Perceived Task Work Load Assessment
- Mental demand: How mentally demanding was the task
- Physical demand: How physically demanding was the task?
- Time demand: How hurried or rushed was the pace of the task?
- Performance: How successful were you in accomplishing what you were asked to do?
- Effort: How hard did you have to work to accomplish your level of performance?
- Frustration level: How insecure, discouraged, irritated, stressed, and annoyed were you

**Objective Understanding Assessment**

1. Consider the following diagram to answer the questions:

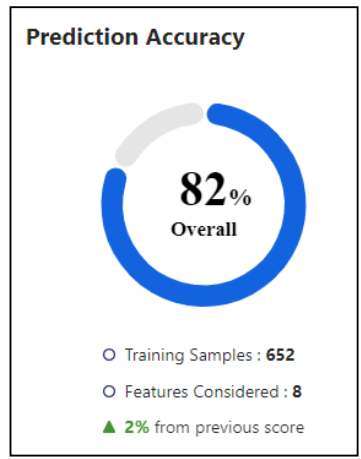

   A. Which of the following statements is **correct**:
      i. I don't know
      ii. The AI system shows the prediction accuracy for the risk of diabetics for a single patient
      iii. **The AI system shows the prediction accuracy for a group of patients who are classified as diabetic or non-diabetic**
      iv. The AI system shows the prediction accuracy for a group of patients who are classified as sick or healthy
   B. Which of the following statements are **correct**:
      i. The current system can only correctly predict the diabetic status of 82 patients
      ii. **The previous prediction accuracy of the system was 80%**
      iii. The current system can only correctly predict for 82 diabetic patients and 82 non-diabetic patients.
      iv. I don't know
   C. Which of the following statements are **correct**:
      i. The current system can only correctly predict the diabetic status of 82 patients
      ii. **If there are 1000 patient records in the database, the current system is expected to predict the diabetes status of 820 patients correctly.**
      iii. The current system can only correctly predict for 82 diabetic patients and 82 non-diabetic patients.
      iv. I don't know

2. Consider the following diagram to answer questions:

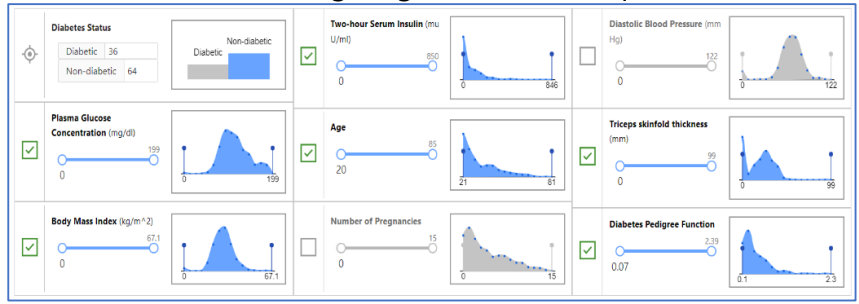

A. How many features have been considered to train the model?
   a. **6**
   b. 7
   c. 8
   d. 9
B. Which of the following features would you like to unselect for improved prediction accuracy?
   a. Age
   b. Number of Pregnancies
   **c. Two-hour Serum Insulin**
   d. Diabetes Pedigree Function
C. Which of the following risk factors seem to have a more symmetrical data distribution?
   a. Two-hour Serum Insulin
   b. **Body Mass Index**
   c. Diabetes Pedigree Function
   d. I don't know

3. Consider the following diagram to answer the question:

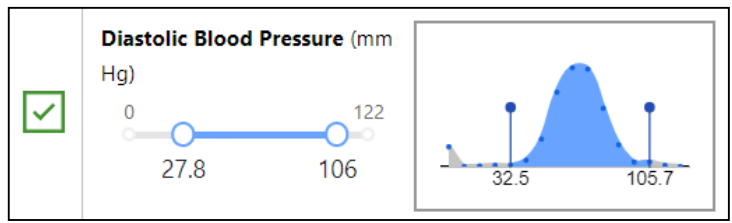

A. Which of the following is **correct**:
   a. Diastolic blood pressure is not considered as a risk factor for training the predictive model
   b. The predictive model will be trained on patient data having a diastolic blood pressure between 0 and 122 mm Hg

c. **The predictive model will be trained on patient data having a diastolic blood pressure between 27.8 and 106 mm Hg**
   d. Diastolic blood pressure of a non-diabetic patient is between 27.8 and 106 mm Hg.

4. Dashboard Specific Questions (These questions will be asked to the specific groups.)

   MCE:

   Consider the following diagram to answer the question:

   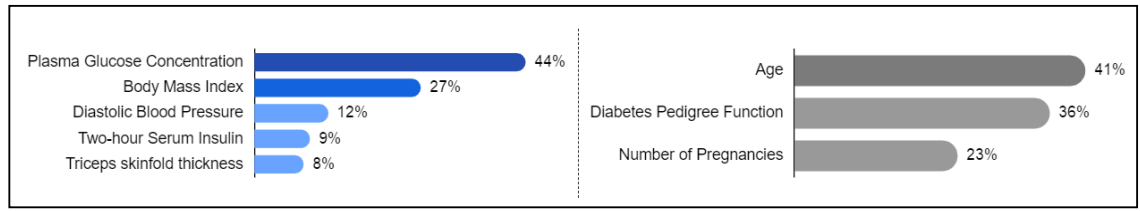

   According to the system, which one is the second most important risk factor which can be controlled effectively by the patients:

   a. Plasma Glucose Concentration
   b. Age
   c. **Body Mass Index**
   d. I don't know

   **(or)**

   DCE:

   Consider the following diagram to answer the question:

   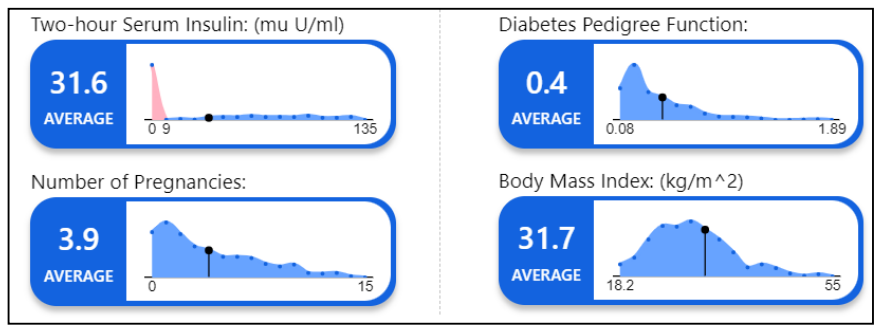

   Which of the following features can negatively affect the prediction accuracy:

   a. **Two-hour Serum Insulin**
   b. Diabetes Pedigree Function
   c. Number of Pregnancies
   d. Body Mass Index

**(or)**

HYB: Will get both 6. MCE and 6. DCE question for 0.5 Mark each

## Perceived Understandability Assessment (Hoffman et al. *)

Q1. I know what will happen the next time I use the system because I understand how it behaves.
Q2. Although I may not know exactly how the system works, I know how to use it to make decisions about the problem.
Q3. It is easy to follow what the system does.
Q4. I recognise what I should do to get the advice I need from the system the next time I use it.

## Perceived Trust Assessment (Cahour-Forzy scale : Hoffman et al. *)

Q1. What is your confidence in the AI system? Do you have a feeling of trust in it?
Q2. Are the actions of the AI system predictable?
Q3. Is the AI system reliable? Do you think it is safe?
Q4. Is the AI system efficient at what it does?

* Robert R. Hoffman, Shane T. Mueller, Gary Klein, and Jordan Litman. 2019. *Metrics for Explainable AI: Challenges and Prospects*

## Study Data and Analysis Scripts

The quantitative data collected from user study 1 and their corresponding analysis Python notebook is provided in the `Study Data and Analysis` folder.

## Non-significant Results From Study 1

1. Objective Understanding

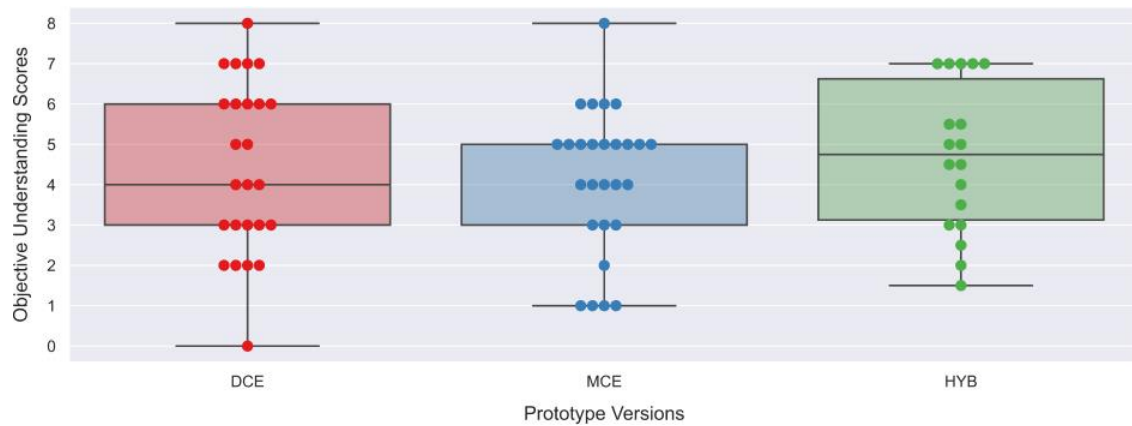

We did not observe any statistically significant differences in the objective understanding scores (H=0.79, p=0.67) using Kruskal-Wallis test.

2. Perceived understanding and trust

We also did not observe any statistically significant differences in perceived understanding and trust.

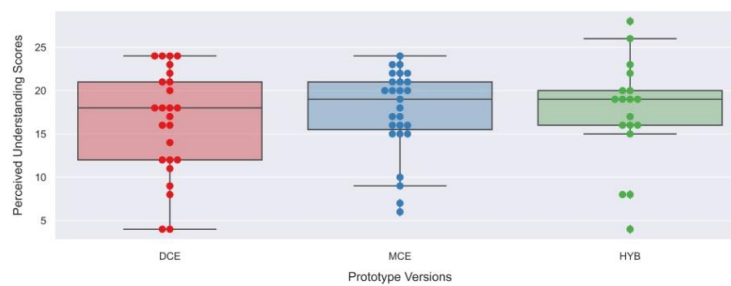

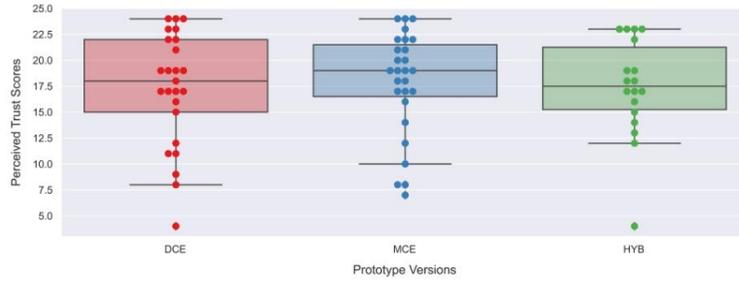

The Kruskal-Wallis test scores for perceived understanding and trust are (H=0.32, p=0.85) and (H=0.33, p=0.85) respectively. The following figure presents diverging bar charts showing the distribution of 7-point Likert Scale responses to perceived understandability and trust

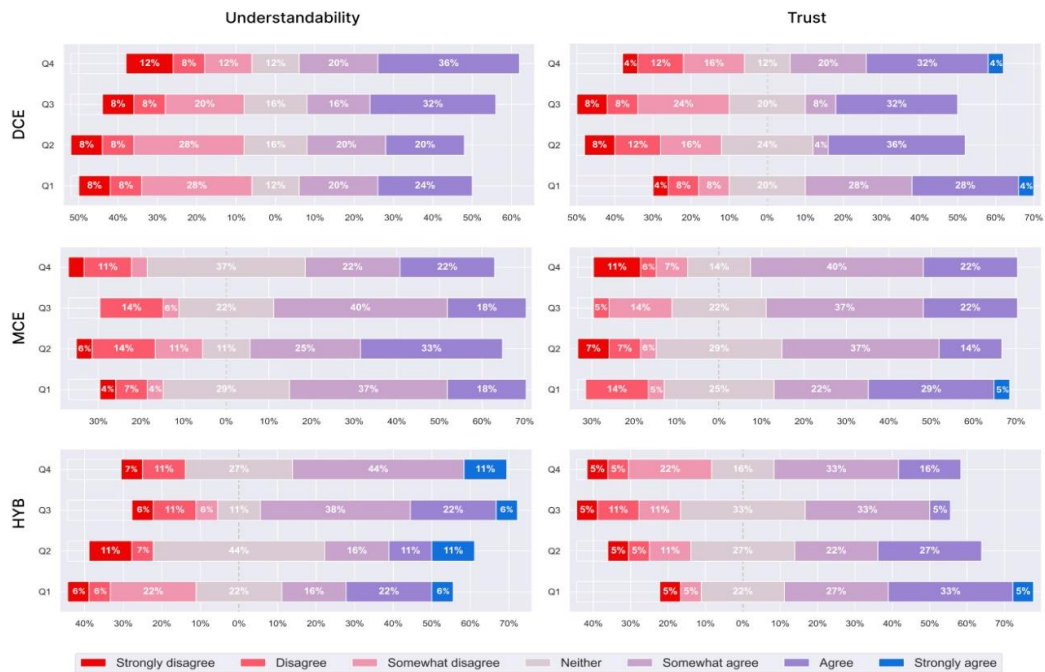

## C. User Study 2: Qualitative Study Protocols and Questions

### Study Introduction
- Informed Consent
- Context Setting
- Prototype Introduction
- Live demonstration of explanation dashboard and data configuration mechanisms
- Self-exploration of prototype

## Demographic Questions
1. Age
2. Gender
3. Occupation
4. Experience in healthcare
5. Field of specialisation

## Semi-Structured Interview Questions

1. What do you understand about the different visuals presented in this explanation dashboard? What is the main information that you can get from each of them?

2. Did you find it easy to understand the explanations?
- Was the information easy/difficult to digest?
- Was it easy/difficult to navigate?

3. What did you learn about the dataset used for training the AI model?

4. Which category/information in the explanations was most helpful for you? Why?

5. Is there something that does not help you, or you felt less important to know?
- What else would you want in the explanations?

6. Can you do anything to increase the prediction accuracy?

7. Out of the manual and automated control mechanisms provided in the system, which one is more helpful to you? Why?

8. Which of the two control mechanisms was least useful to you? Why?

9. If healthcare experts are given the control to improve the ML model, how do you think it can impact their trust in the system? How do you think it can impact their understanding of the system?

9. What do you understand about these data issues?

10. Is it useful if the system provides information about the data issues? Why? How does it help you?

11. If healthcare experts are informed about these data issues, how do you think it can impact their trust in the system? How do you think it can impact their understanding of the system?

12. Do you think these explanations can affect your trust and confidence in the system?


%% \begin{comment}
%% For Supplementary
%% \end{comment}
\end{document}